\documentclass[journal]{IEEEtran}

\usepackage{comment}
\usepackage{cite}
\usepackage{graphicx}
\usepackage{subfigure}
\usepackage{csquotes}
\usepackage{amsfonts}
\usepackage{amsmath}
\usepackage{mathtools}
\usepackage{enumitem}
\usepackage{textcomp}
\usepackage[table]{xcolor}
\usepackage{xspace}
\usepackage{array}
\usepackage{makecell}
\usepackage{scalerel}
\usepackage{multirow}
\usepackage{colortbl}
\usepackage{amssymb}% http://ctan.org/pkg/amssymb
\usepackage{pifont}% http://ctan.org/pkg/pifont
\usepackage{booktabs}
\usepackage{url}
\usepackage{hyperref}
\usepackage{balance} 
\usepackage{mathrsfs}
\usepackage{lipsum}
\usepackage{multicol}
\usepackage{blindtext}
\usepackage{xcolor}

\colorlet{Ra}{green!50!black} 
\colorlet{Rb}{violet!90!black}
\colorlet{Rc}{blue!80}%{red!60!black}
\colorlet{Rall}{magenta!50!red}

\usepackage[para*]{manyfoot}
\DeclareNewFootnote[para]{A}
\let\oldfootnoteA\footnoteA
\renewcommand{\footnoteA}[1]{%
\oldfootnoteA{\makebox[0.5\dimexpr\textwidth-2\footglue\relax][l]{#1}}} % adjust .305\textwidth

\ifCLASSINFOpdf
\else
\fi

\makeatletter
\DeclareRobustCommand{\cev}[1]{%
  {\mathpalette\do@cev{#1}}%
}
\newcommand{\do@cev}[2]{%
  \vbox{\offinterlineskip
    \sbox\z@{$\m@th#1 x$}%
    \ialign{##\cr
      \hidewidth\reflectbox{$\m@th#1\vec{}\mkern4mu$}\hidewidth\cr
      \noalign{\kern-\ht\z@}
      $\m@th#1#2$\cr
    }%
  }%
}

\newcommand{\cmark}{$\bigcirc$}%
\newcommand{\xmark}{\ding{53}}%

\newcommand{\thickhline}{%
    \noalign {\ifnum 0=`}\fi \hrule height 1pt
    \futurelet \reserved@a \@xhline
}

\makeatother

\begin{document}

\title{Learning from Noisy Labels with Deep Neural Networks: A Survey}

\author{Hwanjun Song,
        Minseok Kim,
        Dongmin Park,
        Yooju Shin,
        Jae-Gil Lee% <-this % stops a space
%\thanks{H. Song, M. Kim, D, Park, and J.-G., Lee are with the Graduate School of Knowledge Service Engineering, Korea Advanced Institute of Science and Technology, Daejeon 34141, Republic of Korea (e-mail: songhwanjun@kaist.ac.kr; minseokkim@kaist.ac.kr; dongminpark@kaist.ac.kr; jaegil@kaist.ac.kr).}% <-this % stops a space
\thanks{H. Song is with NAVER AI Lab, Seongnam 13561, Republic of Korea (e-mail: hwanjun.song@navercorp.com);
M. Kim, D, Park, Y, Shin, and J.-G., Lee are with the Graduate School of Knowledge Service Engineering, Korea Advanced Institute of Science and Technology, Daejeon 34141, Republic of Korea (e-mail: minseokkim@kaist.ac.kr; dongminpark@kaist.ac.kr; yooju24@kaist.ac.kr; jaegil@kaist.ac.kr). This work has been submitted to the IEEE for possible publication. Copyright may be transferred without notice, after which this version may no longer be accessible.}
%\thanks{Manuscript received April 5, 2021.}
}


\markboth{IEEE TRANSACTIONS ON NEURAL NETWORKS AND LEARNING SYSTEMS}%
{Song \MakeLowercase{\textit{et al.}}: Learning from Noisy Labels with Deep Neural Networks: A Survey}
% *** Note that you probably will NOT want to include the author's ***https://www.overleaf.com/project/5eda39dcc0498000019ed9e6



\maketitle

\begin{abstract}
Deep learning has achieved remarkable success in numerous domains with help from large amounts of big data. However, the quality of data labels is a concern because of the lack of high-quality labels in many real-world scenarios. As noisy labels severely degrade the generalization performance of deep neural networks, learning from noisy labels\,(robust training) is becoming an important task in modern deep learning applications. In this survey, we first describe the problem of learning with label noise from a supervised learning perspective. Next, we provide a comprehensive review of 62 state-of-the-art robust training methods, all of which are categorized into five groups according to their methodological difference, followed by a systematic comparison of six properties used to evaluate their superiority. Subsequently, we perform an in-depth analysis of noise rate estimation and summarize the typically used evaluation methodology, including public noisy datasets and evaluation metrics. Finally, we present several promising research directions that can serve as a guideline for future studies. All the contents will be available at \href{https://github.com/songhwanjun/Awesome-Noisy-Labels}{\color{blue!50!brown} https://github.com/songhwanjun/Awesome-Noisy-Labels}.
\looseness=-1
\end{abstract}


% \fi
\IEEEpeerreviewmaketitle

%\ifCLASSOPTIONcaptionsoff
%  \newpage
%\fi


\section{Introduction}
\label{sec:introduction}

% whay noisy labels happen in collecting datat?
\IEEEPARstart{W}ith the recent emergence of large-scale datasets, deep neural networks (DNNs) have exhibited impressive performance in numerous machine learning tasks, such as computer vision \cite{krizhevsky2012imagenet,redmon2016you}, information retrieval \cite{zhang2016deep,pang2017deeprank,onal2018neural}, and language processing \cite{howard2018universal,devlin2019bert,severyn2015twitter}. Their success is dependent on the availability of massive but carefully labeled data, which are expensive and time-consuming to obtain. Some non-expert sources, such as Amazon's Mechanical Turk and the surrounding text of collected data, have been widely used to mitigate the high labeling cost; however, the use of these source often results in unreliable labels \cite{paolacci2010running,cothey2004web,mason2012conducting, scott2013classification}. 
In addition, data labels can be extremely complex even for experienced domain experts \cite{frenay2013classification, lloyd2004observer}; 
they can also be adversarially manipulated by a label-flipping attack \cite{xiao2012adversarial}. Such unreliable labels are called \emph{noisy labels} because they may be \emph{corrupted} from ground-truth labels. The ratio of corrupted labels in real-world datasets is reported to range from $8.0\%$ to $38.5\%$ \cite{xiao2015learning,li2017webvision, lee2018cleannet,song2019selfie}. \looseness=-1

\begin{figure}[t!]
\begin{center}
\includegraphics[width=8.8cm]{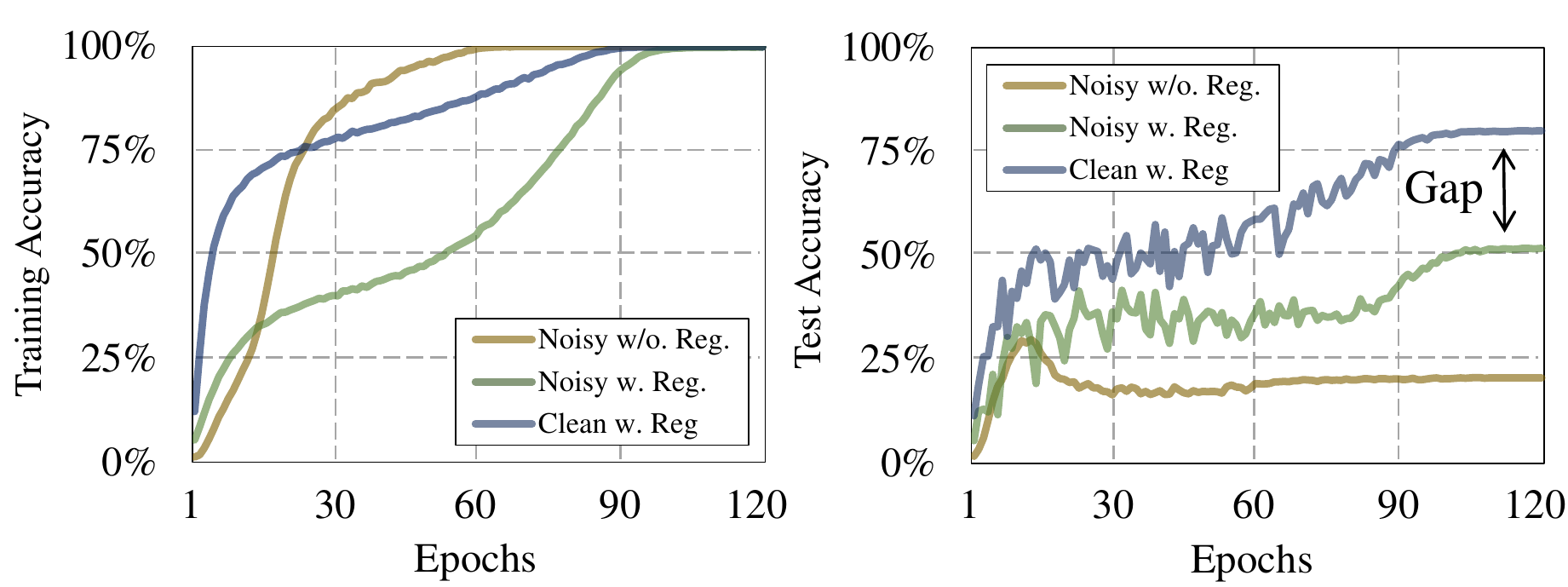}
\end{center}
\vspace*{-0.55cm}
\caption{Convergence curves of training and test accuracy when training WideResNet-16-8 using a standard training method on the CIFAR-100 dataset with the symmetric noise of $40\%$: \enquote{Noisy w/o. Reg.} and \enquote{Noisy w. Reg.} are the models trained on noisy data without and with regularization, respectively, and \enquote{Clean w. Reg.} is the model trained on clean data with regularization.}
\label{fig:convergence_analysis}
\vspace*{-0.4cm}
\end{figure}

In the presence of noisy labels, training DNNs is known to be susceptible to noisy labels because of the significant number of model parameters that render DNNs overfit to even corrupted labels with the capability of learning any complex function \cite{krause2016unreasonable, arpit2017closer}. Zhang et al. \cite{zhang2016understanding} demonstrated that DNNs can easily fit an entire training dataset with any ratio of corrupted labels, which eventually resulted in poor generalizability on a test dataset. Unfortunately, popular regularization techniques, such as data augmentation \cite{shorten2019survey}, weight decay \cite{krogh1992simple}, dropout \cite{srivastava2014dropout}, and batch normalization \cite{ioffe2015batch} {have been applied extensively, but they do \emph{not} completely overcome the overfitting issue by themselves.} As shown in Figure \ref{fig:convergence_analysis}, the gap in test accuracy between models trained on clean and noisy data remains significant even though all of the aforementioned regularization techniques are activated. Additionally, the accuracy drop with label noise is considered to be more harmful than with other noises, such as input noise \cite{zhu2004class}. Hence, achieving a good generalization capability in the presence of noisy labels is a key challenge.

% recent noisy label for deep learning
Several studies have been conducted to investigate supervised learning under noisy labels. Beyond conventional machine learning techniques \cite{frenay2013classification, zhang2016learning}, deep learning techniques have recently gained significant attention in the machine learning community. In this survey, we present the advances in recent deep learning techniques for overcoming noisy labels. We surveyed {recent studies by recursively tracking relevant bibliographies in papers published at premier research conferences, such as CVPR, ICCV, NeurIPS, ICML, and ICLR. Although we attempted to comprehensively include all recent studies at the time of submission, some of them may not be included because of the quadratic increase in deep learning papers. The studies included were grouped into \emph{five} categories, as shown in Figure \ref{fig:categorization} (see Section \ref{sec:methodology} for details).

\begin{figure*}[t!]
\begin{center}
\includegraphics[width=16.5cm]{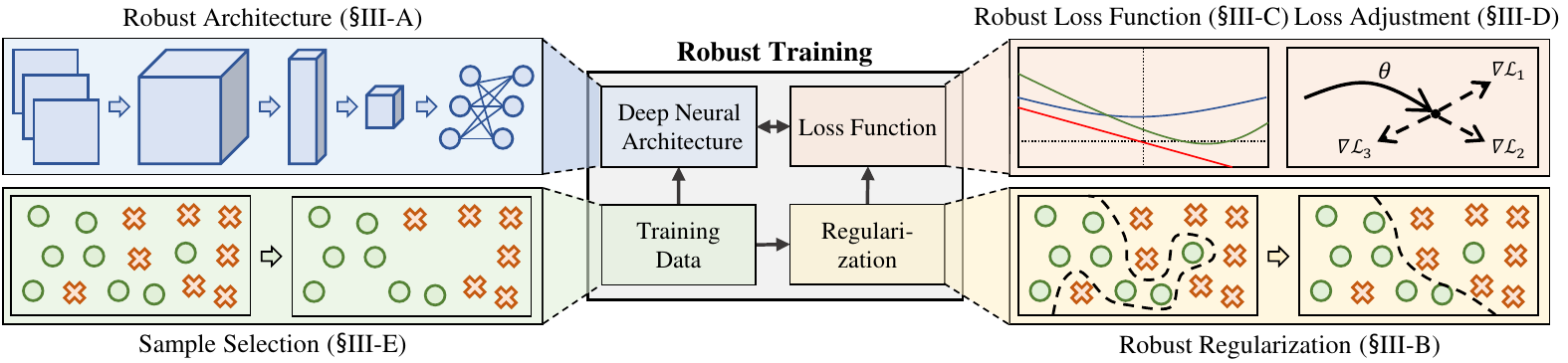}
\end{center}
\vspace*{-0.55cm}
\caption{Categorization of recent deep learning methods for overcomming noisy labels.}
\label{fig:categorization}
\vspace*{-0.4cm}
\end{figure*}

%%%%%%%%%%%%%%%%%%%%%%%%%%%%%%%%%%%%%%%%%%%%%%%%%%%%%%%%%%%%%%%%%%%%%%%%%%%%%%%%%%%%%%%%%%%%
% put the organization of this paper here!

%%%%%%%%%%%%%%%%%%%%%%%%%%%%%%%%%%%%%%%%%%%%%%%%%%%%%%%%%%%%%%%%%%%%%%%%%%%%%%%%%%%%%%%%%%%%

\subsection{Related Surveys}
\label{sec:related_surveys}

Fr{\'e}nay and Verleysen \cite{frenay2013classification} discussed the potential negative consequence of learning from noisy labels and provided a comprehensive survey on noise-robust classification methods, focusing on conventional supervised approaches such as na\"ive Bayes and support vector machines. Furthermore, their survey included the definitions and sources of label noise as well as the taxonomy of label noise. Zhang et al. \cite{zhang2016learning} discussed another aspect of label noise in crowdsourced data annotated by non-experts and provided a thorough review of expectation-maximization (EM) algorithms that were proposed to improve the quality of crowdsourced labels. Meanwhile, Nigam et al. \cite{nigam2020impact} provided a brief introduction to deep learning algorithms that were proposed to manage noisy labels; however, the scope of these algorithms was limited to only two categories, i.e., the loss function and sample selection in Figure \ref{fig:categorization}. Recently, Han et al. \cite{han2020survey} summarized the essential components of robust learning with noisy labels, but their categorization is totally different from ours in philosophy; {we mainly focus on systematic methodological difference, whereas they rather focused on more general views, such as input data, objective functions, and optimization policies. Furthermore, this survey is the first to present a comprehensive methodological comparison of existing robust training approaches (see Tables \ref{table:all_comparision} and \ref{table:direction_comparison}}).  %Furthermore, we presents a comprehensive survey based on the systematic and clear taxonomy at a level of technical depth.}

\subsection{Survey Scope}
\label{sec:survey_scope}

{
Robust training with DNNs becomes critical to guarantee the reliability of machine learning algorithms. In addition to label noise, two types of flawed training data have been actively studied by different communities \cite{akhtar2018threat, yoon2018gain}. \emph{Adversarial learning} is designed for small, worst-case perturbations of the inputs, so-called adversarial examples, which are maliciously constructed to deceive an already trained model into making errors \cite{fawzi2016robustness, dohmatob2018limitations, gilmer2019adversarial, mahloujifar2019curse}. Meanwhile, \emph{data imputation} primarily deals with missing inputs in training data, where missing values are estimated from the observed ones \cite{rubin1976inference, yoon2018gain}. Adversarial learning and data imputation are closely related to robust learning, but handling \emph{feature} noise is beyond the scope of this survey---i.e., learning from noisy \emph{labels}. 
}
\section{Preliminaries}
\label{sec:preliminaries}

In this section, the problem statement for supervised learning with noisy labels is provided along with the taxonomy of label noise. Managing noisy labels is a long-standing issue; therefore, we review the basic conventional approaches and {theoretical foundations underlying robust deep learning}. Table \ref{table:notation} summarizes the notation frequently used in this study. \looseness=-1

{
\newcolumntype{L}[1]{>{\centering\let\newline\\\arraybackslash\hspace{0pt}}m{#1}}
\newcolumntype{X}[1]{>{\let\newline\\\arraybackslash\hspace{0pt}}p{#1}}
\begin{table}[t!]
\caption{Summary of the notation.}
\vspace*{-0.3cm}
\begin{center}
\begin{tabular}{|L{1.7cm} |X{6.2cm}|}\hline
\!\!\textbf{Notation} & \hspace*{2.25cm} \textbf{Description}  \\\hline \hline 
$\mathcal{X}$ & the data feature space \\\hline
$\mathcal{Y}$, $\tilde{\mathcal{Y}}$ & the true and noisy label space \\\hline
$\mathcal{D}$, $\tilde{\mathcal{D}}$ & the clean and noisy training data \\\hline
$P_{\mathcal{D}}$, $P_{\tilde{\mathcal{D}}}$ & the joint distributions of clean and noisy data \\\hline
$\mathcal{B}_{t}$ & a set of mini-batch examples at time $t$ \\\hline
$\Theta_{t}$ & the parameter of a deep neural network at time $t$ \\\hline
$f(\,\cdot\,;\Theta_{t})$ & a deep neural network parameterized by $\Theta_{t}$ \\\hline
$\ell$ & a specific loss function \\\hline
$\mathcal{R}$ & an empirical risk \\\hline
$\mathbb{E}_{\mathcal{D}}$ & an expectation over $\mathcal{D}$ \\\hline
$x$, $x_i$ & a data example of $\mathcal{X}$ \\\hline
$y$, $y_i$ & a true label of $\mathcal{Y}$ \\\hline
$\tilde{y}$, $\tilde{y}_i$ & a noisy label of $\tilde{\mathcal{Y}}$ \\\hline
$\eta$ & a specific learning rate \\\hline
$\tau$ & a true noise rate \\\hline
$b$ & the number of mini-batch examples in $\mathcal{B}_{t}$ \\\hline
$c$ & the number of classes \\\hline
{T}, $\hat{\text{T}}$ & the true and estimated noise transition matrix \\\hline
\end{tabular}
\end{center}
\label{table:notation}
\vspace*{-0.55cm}
\end{table}
}

\begin{figure*}[t!]
\begin{center}
\includegraphics[width=15.5cm]{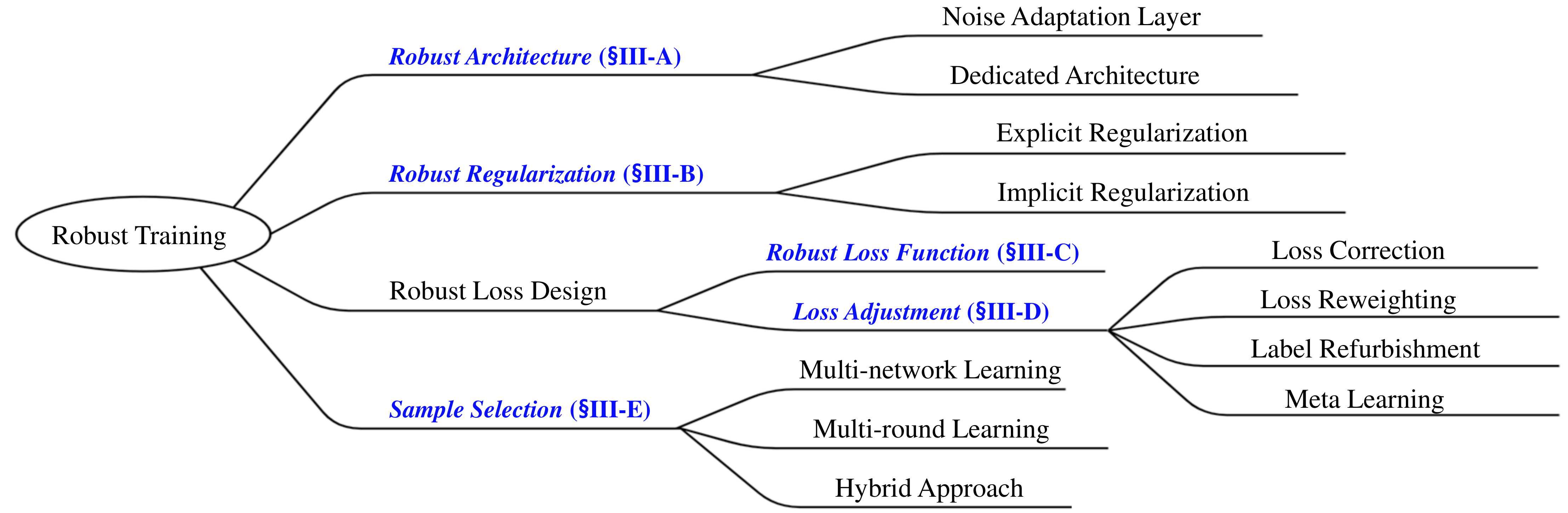}
\end{center}
\vspace*{-0.45cm}
\caption{A high level research overview of robust deep learning for noisy labels. The research directions that are actively contributed by the machine learning community are categorized into five groups in blue italic.}
\label{fig:tree_categorization}
\vspace*{-0.4cm}
\end{figure*}

%\vspace*{-0.1cm}
\subsection{Supervised Learning with Noisy Labels}
%\vspace*{-0.1cm}

\emph{Classification} is a representative supervised learning task for learning a function that maps an input feature to a label \cite{bishop2006pattern}. In this paper, we consider a $c$-class classification problem using a DNN with a softmax output layer. Let $\mathcal{X} \subset \mathbb{R}^{d}$ be the feature space and $\mathcal{Y} = \{0,1\}^{c}$ be the ground-truth label space in a \emph{one-hot} manner. In a typical classification problem, we are provided with a training dataset $\mathcal{D}=\{(x_i, y_i)\}_{i=1}^{N}$ obtained from an unknown joint distribution $P_{\mathcal{D}}$ over $\mathcal{X} \times \mathcal{Y}$, where each $(x_i, y_i)$ is \emph{independent and identically distributed}. The goal of the task is to learn the mapping function $f(\,\cdot\,;\Theta): \mathcal{X} \rightarrow [0, 1]^{c}$ of the DNN parameterized by $\Theta$ such that the parameter $\Theta$ minimizes the empirical risk $\mathcal{R}_{\mathcal{D}}(f)$,
\begin{equation}
\label{eq:empirical_risk}
\!\!\mathcal{R}_{\mathcal{D}}(f) = \mathbb{E}_{\mathcal{D}}[\ell\big(f(x;\Theta), y\big)] = \frac{1}{|\mathcal{D}|}\!\sum_{(x,y)\in\mathcal{D}}\!\!\!\!\ell\big(f(x;\Theta), y\big),
\end{equation}
where $\ell$ is a certain loss function.

As data labels are corrupted in various real-world scenarios, we aim to train the DNN from noisy labels. Specifically, we are provided with a noisy training dataset $\tilde{\mathcal{D}}=\{(x_i, \tilde{y}_i)\}_{i=1}^{N}$ obtained from a noisy joint distribution $P_{\tilde{\mathcal{D}}}$ over $\mathcal{X}\times\tilde{\mathcal{Y}}$, where $\tilde{y}$ is a \emph{noisy} label which may not be true. Hence, following the standard training procedure, a mini-batch $\mathcal{B}_{t}=\{(x_i, \tilde{y}_i)\}_{i=1}^{b}$ comprising $b$ examples is obtained randomly from the noisy training dataset $\tilde{\mathcal{D}}$ at time $t$. Subsequently, the DNN parameter $\Theta_t$ at time $t$ is updated along the descent direction of the empirical risk on mini-batch $\mathcal{B}_t$,
\begin{equation}
\label{eq:corrupted_update}
\Theta_{t+1} = \Theta_{t} - \eta\nabla\Big(\frac{1}{|\mathcal{B}_{t}|} \!\sum_{(x,\tilde{y}) \in \mathcal{B}_{t}} \!\!\!\!\ell\big(f(x;\Theta_{t}), \tilde{y}\big)\Big),
\end{equation}
where $\eta$ is a learning rate specified. \looseness=-1

Here, the risk minimization process is no longer \emph{noise-tolerant} because of the loss computed by the noisy labels. DNNs can easily memorize corrupted labels and correspondingly degenerate their generalizations on unseen data \cite{frenay2013classification, zhang2016learning, nigam2020impact}. Hence,  mitigating the adverse effects of noisy labels is essential to enable noise-tolerant training for deep learning. %\looseness=-1
%\jaegil{deep neural network, network $\rightarrow$ DNN (use a consistent terminology)}

\vspace*{-0.1cm}
\subsection{Taxonomy of Label Noise}
\label{sec:taxonomy}

{This section presents the types of label noise that have been adopted to design robust training algorithms.} Even if data labels are corrupted from ground-truth labels without \emph{any} prior assumption, in essence, the corruption probability is affected by the dependency between \emph{data features} or \emph{class labels}. A detailed analysis of the taxonomy of label noise was provided by Fr{\'e}nay and Verleysen \cite{frenay2013classification}. {Most existing algorithms dealt with instance-independent noise, but instance-dependent noise has not yet been extensively investigated owing to its complex modeling.}

\smallskip
\subsubsection{\textbf{Instance-independent Label Noise}}
A typical approach for modeling label noise assumes that the corruption process is conditionally \emph{independent} of data features when the true label is given \cite{natarajan2013learning,zhang2016understanding}. That is, the true label is corrupted by a \emph{noise transition matrix} $\text{T} \in [0, 1]^{c\times c}$, where $ \text{T}_{ij} \coloneqq p(\tilde{y}=j|y=i)$ is the probability of the true label $i$ being flipped into a corrupted label $j$. 
In this approach, the noise is called a \emph{symmetric}\,(or \emph{uniform}) noise with a noise rate $\tau \in [0,1]$ if $\forall_{i=j} \text{T}_{ij} \!=\! 1-\tau \wedge \forall_{i \neq j} \text{T}_{ij} = \frac{\tau}{c-1}$, where a true label is flipped into other labels with equal probability. In contrast to symmetric noise, the noise is called an \emph{asymmetric}\,(or \emph{label-dependent}) noise if $\forall_{i=j} \text{T}_{ij} \!=\! 1-\tau \wedge \exists_{i \neq j, i\neq k, j\neq k} \text{T}_{ij} > \text{T}_{ik} $, where a true label is more likely to be mislabeled into a particular label. For example, a \enquote{dog} is more likely to be confused with a \enquote{cat} than with a \enquote{fish.} In a stricter case when $\forall_{i=j} \text{T}_{ij} \!=\! 1-\tau \wedge \exists_{i \neq j} \text{T}_{ij} = \tau$, the noise is called a \emph{pair noise}, where a true label is flipped into only a certain label.  

\smallskip
\subsubsection{\textbf{Instance-dependent Label Noise}}
For more realistic noise modeling, the corruption probability is assumed to be \emph{dependent} on both the data features and class labels \cite{xiao2015learning, goldberger2016training}. Accordingly, the corruption probability is defined as $\rho_{ij}(x)\!= \!p(\tilde{y}\!=\!j|y\!=\!i, x)$. 
%Unlike the aforementioned noises, because the data feature of an example $x$ also affects the chance of $x$ being mislabeled, the noise is called an \emph{instance-dependent} noise. 
Unlike the aforementioned noises, the data feature of an example $x$ also affects the chance of $x$ being mislabeled. %However, the modeling of this noise has not been investigated extensively yet owing to its complexity. 

\vspace*{-0.1cm}
\subsection{Non-deep Learning Approaches}
\label{sec:non_deep_learning}

For decades, numerous methods have been proposed to manage noisy labels using conventional machine learning techniques. These methods can be categorized into \emph{four} groups \cite{frenay2013classification, sastry2017robust, nigam2020impact}, as follows:

\begin{itemize}[leftmargin=9pt]
\item \textbf{Data Cleaning:} Training data are cleaned by excluding examples whose labels are likely to be corrupted. Bagging and boosting are used to filter out false-labeled examples to remove examples with higher weights because false-labeled examples tend to exhibit much higher weights than true-labeled examples \cite{wheway2000using, sluban2014ensemble}. In addition, various methods, such as $k$-nearest neighbor, outlier detection, and anomaly detection, have been widely exploited to exclude false-labeled examples from noisy training data \cite{delany2012profiling, gamberger2000noise, thongkam2008support}. Nevertheless, this family of methods suffers from over-cleaning issue that overly removes even the true-labeled examples. %\looseness=-1 % \cite{frenay2013classification}.
\vspace*{0.12cm}
\item \textbf{Surrogate Loss:} Motivated by the noise-tolerance of the 0-1 loss function \cite{natarajan2013learning}, many researchers have attempted to resolve its inherent limitations, such as computational hardness and non-convexity  that render gradient methods unusable. Hence, several convex surrogate loss functions, which approximate the 0-1 loss function, have been proposed to train a specified classifier under the binary classification setting \cite{mnih2012learning, manwani2013noise, ghosh2015making, van2015learning, patrini2016loss}. However, these loss functions cannot support the multi-class classification task.
% bagging and boosting
\vspace*{0.12cm}
\item \textbf{Probabilistic Method:} Under the assumption that the distribution of features is helpful in solving the problem of learning from noisy labels \cite{xu2005survey}, the confidence of each label is estimated by clustering and then used for a weighted training scheme \cite{rebbapragada2007class}. This confidence is also used to convert hard labels into soft labels to reflect the uncertainty of labels \cite{liu2017soft}. In addition to these clustering approaches, several Bayesian methods have been proposed for graphical models such that they can benefit from using any type of prior information in the learning process \cite{kaster2010comparative}. However, this family of methods may exacerbate the overfitting issue owing to the increased number of model parameters.  
\vspace*{0.12cm}
\item \textbf{Model-based Method:} As conventional models, such as the SVM and decision tree, are not robust to noisy labels, significant effort has been expended to improve the robustness of them. To develop a robust SVM model, misclassified examples during learning are penalized in the objective \cite{ganapathiraju2000support, biggio2011support}. In addition, several decision tree models are extended using new split criteria to solve the overfitting issue when the training data are not fully reliable \cite{mantas2014credal, ghosh2017robustness}. However, it is infeasible to apply their design principles to deep learning.
\end{itemize}

\smallskip
{
Meanwhile, deep learning is more susceptible to label noises than traditional machine learning owing to its high expressive power, as proven by many researchers \cite{arpit2017closer, liu2020early, li2020gradient}. 
There has been significant effort to understand why noisy labels negatively affect the performance of DNNs \cite{patrini2017making, zhang2016understanding, li2020gradient, cheng2020learning}. This theoretical understanding has led to the algorithmic design which achieves higher robustness than non-deep learning methods. A detailed analysis of theoretical understanding for robust deep learning was provided by Han et al. \cite{han2020survey}. 
}

\vspace*{-0.2cm}
\subsection{Regression with Noisy Labels}
\label{sec:regression}

{
In addition to classification, regression is another main topic of supervised machine learning, which aims to model the relationship between a number of features and a continuous target variable. Unlike the classification task with a \emph{discrete} label space, the regression task considers the continuous variable as its target label \cite{garg2020robust}, and thus it learns the mapping function $f(~\cdot~; \Theta): \mathcal{X} \rightarrow \mathcal{Y}$, where $\mathcal{Y} \in \mathbb{R}$ is a \emph{continuous} label space. 
Given the input feature $x$ and its ground-truth label $y$, two types of label noise are considered in the regression task. An \emph{additive noise} \cite{hu2019simple} is formulated by $\tilde{y} := y + \epsilon$ where $\epsilon$ is drawn from a random distribution  independent from the input feature; an \emph{instance-dependent noise} \cite{menon2018learning} is formulated by $\tilde{y} := \rho(x)$ where $\rho: \mathcal{X} \rightarrow \mathcal{Y}$ is a noise function dependent on the input feature.  
%Although these label noises are quite different compared with symmetric and asymmetric noises for the classification task, some of algorithms in Table \ref{table:all_comparision} can be used for both classification and regression with small modifications if discrete labels are not necessary.
%and  However, other directions cannot easily applied to the regression task because discrete labels are explicitly required.

Although regression predicts continuous values, regression and classification share the same concept of learning the mapping function from the input feature $x$ to the output label $y$. Thus, many robust approaches for classification are easily extended to the regression problem with simple modification \cite{torgo1997regression}. Thus, in this survey, we focus on the classification setting for which most robust methods are defined.
}
\vspace*{-0.1cm}
\section{Deep Learning Approaches}
\label{sec:methodology}
\vspace*{-0.0cm}

According to our comprehensive survey, the robustness of deep learning can be enhanced in numerous approaches \cite{ghosh2017robust, xiao2015learning, srivastava2014dropout, reed2015training, malach2017decoupling, garcia2016noise, yan2016robust, harutyunyan2020improving, chen2021noise}. Figure \ref{fig:tree_categorization} shows an overview of recent research directions conducted by the machine learning community. {All of them\,(i.e., \textsection \ref{sec:robust_architecture}~--~\textsection \ref{sec:sample_selection}) focused on making a supervised learning process more robust to label noise: \looseness=-1

\begin{itemize}[leftmargin=9pt]
\item
(\textsection \ref{sec:robust_architecture}) Robust architecture: adding a noise adaptation layer at the top of an underlying DNN to learn label transition process or developing a dedicated architecture to reliably support more diverse types of label noise;
\vspace*{0.12cm}
\item (\textsection \ref{sec:robust_regularization}) Robust regularization: enforcing a DNN to overfit less to false-labeled examples explicitly or implicitly;
\vspace*{0.12cm}
\item (\textsection \ref{sec:robust_loss_function}) Robust loss function: improving the loss function;
\vspace*{0.12cm}
\item (\textsection \ref{sec:loss_adjustment}) Loss adjustment: adjusting the loss value according to the confidence of a given loss (or label) by loss correction, loss reweighting, or label refurbishment; 
\vspace*{0.12cm}
\item (\textsection \ref{sec:sample_selection}) Sample selection: identifying true-labeled examples from noisy training data via multi-network or multi-round learning. 
%which are also combined with semi-supervised learning to achieve the state-of-the-art performance.
\end{itemize}

%Beyond supervised learning, researchers have recently attempted to further improve noise robustness by adopting \emph{meta learning}\,(\textsection \ref{sec:meta_learning}) and \emph{semi-supervised learning}\,(\textsection \ref{sec:semi_supervised_learning}).
Overall, we categorize all recent deep learning methods into \emph{five} groups corresponding to popular research directions, as shown in Figure \ref{fig:tree_categorization}. In \textsection \ref{sec:loss_adjustment}, meta learning is also discussed because it finds the optimal hyperparameters for loss reweighting. In \textsection \ref{sec:sample_selection}, we discuss the recent efforts for combining sample selection with other orthogonal directions or semi-supervised learning toward the state-of-the-art performance.}

Figure \ref{fig:categorization} illustrates the categorization of robust training methods using these five groups. % Table \ref{table:all_comparision} summarizes existing deep learning methods according to them, which also provides a comparative analysis. %Some methods may belong to more than one categories if they combine multiple approaches. 

\subsection{Robust Architecture}
\label{sec:robust_architecture}

In numerous studies, architectural changes have been made to model the noise transition matrix of a noisy dataset \cite{chen2015webly, bekker2016training, sukhbaatar2014training, jindal2016learning, goldberger2017training, xiao2015learning, han2018masking, yao2018deep, cheng2020weakly}. These changes include adding a noise adaptation layer at the top of the softmax layer and designing a new dedicated architecture. The resulting architectures yield improved generalization through the modification of the DNN output based on the estimated label transition probability.% 

\vspace*{0.15cm}
\subsubsection{\textbf{Noise Adaptation Layer}}
\label{sec:adaptation_layer}

{
From the view of training data, the noise process is modeled by discovering the underlying label transition pattern (i.e., the {noise transition matrix} T). Given an example $x$, the noisy class posterior probability for an example $x$ is expressed by
\begin{equation}
\label{eq:label_transition_matrix}
\begin{gathered}
\!\!\!\!\!p(\tilde{y}=j|x) \!=\!\sum_{i=1}^{c}p(\tilde{y}=j, y=i|x)\! = \!\sum_{i=1}^{c}\text{T}_{ij}p(y=i|x),\\
\text{where} ~~ \text{T}_{ij} = p(\tilde{y}=j|y=i,x).
\end{gathered}
\end{equation} %By designing an accurate estimator of label transition matri
%Thus, the notion of the noise transition matrix has been employed to build robust approaches via a noise adaptation layer \cite{sukhbaatar2014training, srivastava2014dropout, goldberger2017training} or loss correction \cite{patrini2017making, hendrycks2018using}, both of which aim to calibrate the output of corrupted DNNs.

In light of this, the noise adaptation layer is intended to mimic the label transition behavior in learning a DNN. Let $p(y|x;\Theta)$ be the output of the base DNN with a softmax output layer.
Then, following  Eq.\,\eqref{eq:label_transition_matrix}, the probability of an example $x$ being predicted as its noisy label $\tilde{y}$ is parameterized by }
\begin{equation}
\label{eq:noise_adaption_process}
\begin{split}
\!\!\!p(\tilde{y}=j|x;\Theta, \mathcal{W})&= \sum_{i=1}^{c}p(\tilde{y}=j, y\!=\!i|x; \Theta, \mathcal{W})\\
&= \sum_{i=1}^{c}\underbrace{p(\tilde{y}=j|y\!=\!i;\mathcal{W})}_{\text{Noise Adaptation Layer}}\underbrace{p(y\!=\!i|x;\Theta)}_{\text{Base Model}}.
\end{split}
\end{equation}
Here, the noisy label $\tilde{y}$ is assumed to be \emph{conditionally independent} of the input $x$ in general. 
Accordingly, as shown in Figure \ref{fig:noise_adaption_process}, the noisy adaptation layer is added at the top of the base DNN to model the noise transition matrix parameterized by $\mathcal{W}$. This layer should be removed when test data is to be predicted. \looseness=-1

\begin{figure}[t!]
\begin{center}
\includegraphics[width=8.5cm]{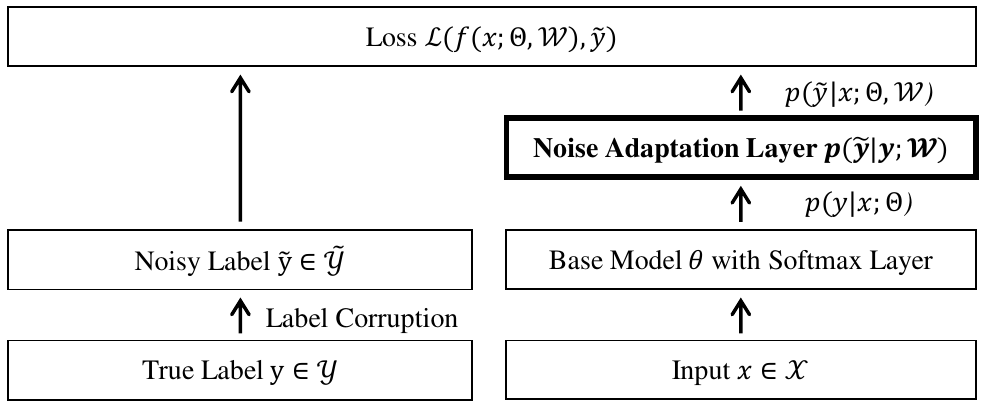}
\end{center}
\vspace*{-0.4cm}
\caption{Noise modeling process using the noise adaptation layer.}
\label{fig:noise_adaption_process}
\vspace*{-0.3cm}
\end{figure}

\smallskip
\noindent \underline{Technical Detail}: \emph{Webly learning} \cite{chen2015webly} first trains the base DNN only for easy examples retrieved by search engines; subsequently, the confusion matrix for all training examples is used as the initial weight $\mathcal{W}$ of the noise adaptation layer. It fine-tunes the entire model in an end-to-end manner for hard training examples. In contrast, the \emph{noise model} \cite{sukhbaatar2014training} initializes $\mathcal{W}$ to an identity matrix and adds a regularizer to force $\mathcal{W}$ to diffuse during DNN training. The \emph{dropout noise model} \cite{srivastava2014dropout} applies dropout regularization to the adaptation layer, whose output is normalized by the softmax function to implicitly diffuse $\mathcal{W}$. The \emph{s-model} \cite{goldberger2017training} is similar to the \emph{dropout noise model} but dropout is not applied. The \emph{c-model} \cite{goldberger2017training} is an extension of the s-model that models the instance-dependent noise, which is more realistic than the symmetric and asymmetric noises. Meanwhile, \emph{NLNN} \cite{bekker2016training} adopts the EM algorithm to iterate the E-step to estimate the noise transition matrix and the M-step to back-propagate the DNN. \looseness=-1

\smallskip
\noindent {\underline{Remark}: A common drawback of this family is their inability to identify false-labeled examples, treating all the examples equally. Thus, the estimation error for the transition matrix is generally large when only noisy training data is used or when the noise rate is high \cite{xia2020extended}.}
%This family is the strong assumption regarding the noise type, which hinders a model's generalization to complex label noise \cite{xiao2015learning}. 
Meanwhile, for the EM-based method, becoming stuck in local optima is inevitable, and high computational costs are incurred \cite{goldberger2017training}.

\vspace*{0.15cm}
\subsubsection{\textbf{Dedicated Architecture}} 
%To overcome the aforementioned drawbacks of the noise adaptation layer, 
{Beyond the label-dependent label noise, several studies have been conducted to support more complex noise, leading to the design of dedicated architectures \cite{xiao2015learning, han2018masking, yao2018deep}.} They typically aimed at increasing the reliability of estimating the label transition probability to handle more complex and realistic label noise. 

%In particular, 
\smallskip
\noindent \underline{Technical Detail}: \emph{Probabilistic noise modeling} \cite{xiao2015learning} manages two independent networks, each of which is specialized to predict the noise type and label transition probability.
Because an EM-based approach with random initialization is impractical for training the entire network, both networks are trained with massive noisy labeled data after the pre-training step with a small amount of clean data.
Meanwhile, \emph{masking} \cite{han2018masking} is a human-assisted approach to convey the human cognition of invalid label transitions.
Considering that noisy labels are mainly from the interaction between humans and tasks, the invalid transition investigated by humans was leveraged to constrain the noise modeling process. Owing to the difficulty in specifying the explicit constraint, a variant of generative adversarial networks\,(GANs) \cite{goodfellow2014generative} was employed in this study. Recently, the \emph{contrastive-additive noise network} \cite{yao2018deep} was proposed to adjust incorrectly estimated label transition probabilities by introducing a new concept of quality embedding, which models the trustworthiness of noisy labels. {\emph{RoG} \cite{lee2019robust} builds a simple yet robust generative classifier on top of any discriminative DNN pre-trained on noisy data.}
  
\smallskip
\noindent \underline{Remark}: Compared with the noise adaptation layer, this family of methods significantly improves the robustness to more diverse types of label noise, but it cannot be easily extended to other architectures in general. 

\subsection{Robust Regularization}
\label{sec:robust_regularization}

Regularization methods have been widely studied to improve the generalizability of a learned model in the machine learning community \cite{ioffe2015batch, krogh1992simple, srivastava2014dropout, shorten2019survey}. 
% They have focused on making slight modifications to the learning algorithm such that the model generalizes better to unseen data.
By avoiding overfitting in model training, the robustness to label noise improves with widely-used regularization techniques such as \emph{data augmentation} \cite{shorten2019survey}, \emph{weight decay} \cite{krogh1992simple}, \emph{dropout} \cite{srivastava2014dropout}, and \emph{batch normalization} \cite{ioffe2015batch}. {These canonical regularization methods operate well on moderately noisy data, but they alone do \emph{not sufficiently} improve the test accuracy}; poor generalization could be obtained when the noise is heavy \cite{tanno2019learning}. Thus, more advanced regularization techniques have been recently proposed, which further improved robustness to label noise when used along with the canonical methods. The main advantage of this family is its \emph{flexibility} in collaborating with other directions because it only requires simple modifications. 

\smallskip
\subsubsection{\textbf{Explicit Regularization}}
{
The regularization can be an {explicit} form that modifies the expected training loss, e.g., weight decay and dropout. 
}

\smallskip
\noindent \underline{Technical Detail}:
\emph{Bilevel learning} \cite{jenni2018deep} uses a clean validation dataset to regularize the overfitting of a model by introducing a bilevel optimization approach, which differs from the conventional one in that its regularization constraint is also an optimization problem. Overfitting is controlled by adjusting the weights on each mini-batch and selecting their values such that they minimize the error on the validation dataset. Meanwhile, \emph{annotator confusion} \cite{tanno2019learning} assumes the existence of multiple annotators and introduces a regularized EM-based approach to model the label transition probability; its regularizer enables the estimated transition probability to converge to the true confusion matrix of the annotators. 
In contrast, \emph{pre-training} \cite{hendrycks2019using} empirically proves that fine-tuning on a pre-trained model provides a significant improvement in robustness compared with models trained from scratch; the universal representations of pre-training prevent the model parameters from being updated in the wrong direction by noisy labels. { \emph{PHuber} \cite{menon2020can} proposes a composite loss-based gradient clipping, which is a variation of standard gradient clipping for label noise robustness. \emph{Robust early-learning} \cite{xia2021robust} classifies critical parameters and non-critical parameters for fitting clean and noise labels, respectively. Then, it penalizes only the non-critical ones with a different update rule.} 
{\emph{ODLN} \cite{wei2021open} leverages open-set auxiliary data and prevents the overfitting to noisy labels by assigning random labels to the open-set examples, which are uniformly sampled from the label set.}

\smallskip
\noindent {\underline{Remark}: The explicit regularization often introduces sensitive model-dependent hyperparameters or requires deeper architectures to compensate for the reduced capacity, yet it can lead to significant performance gain if they are optimally tuned. \looseness=-1}

\smallskip
\subsubsection{\textbf{Implicit Regularization}}
The regularization can also be an {implicit} form that gives the effect of stochasticity, e.g., data augmentation and mini-batch stochastic gradient descent.

\smallskip
\noindent \underline{Technical Detail}: \emph{Adversarial training} \cite{goodfellow2014explaining} enhances the noise tolerance by encouraging the DNN to correctly classify both original inputs and hostilely perturbed ones. {\emph{Label smoothing} \cite{pereyra2017regularizing, lukasik2020does} estimates the marginalized effect of label noise during training, thereby reducing overfitting by preventing the DNN from assigning a full probability to noisy training examples. Instead of the one-hot label, the noisy label is mixed with a uniform mixture over all possible labels,
\begin{equation}
\begin{gathered}
\bar{y} = \big\langle \bar{y}(1), \bar{y}(2), \dots, \bar{y}(c) \big\rangle,\\ 
\text{where} ~ \bar{y}(i) = (1-\alpha) \cdot [\tilde{y} = i] + \alpha / c ~\text{and}~ \alpha \in [0, 1].
\end{gathered}
\end{equation}
Here, $[\cdot]$ is the Iverson bracket and $\alpha$ is the smoothing degree.} In contrast, \emph{mixup} \cite{zhang2018mixup} regularizes the DNN to favor simple linear behaviors in between training examples. 
% Compared to the common data augmentation, mixup models the vicinity relation across the samples of different classes.
First, the mini-batch is constructed using virtual training examples, each of which is formed by the linear interpolation of two noisy training examples $(x_i, \tilde{y}_i)$ and $(x_j, \tilde{y}_j)$ obtained at random from noisy training data $\tilde{\mathcal{D}}$,
\begin{equation}
\label{eq:mixup_construction}
{x}_{mix} = \lambda x_i + (1-\lambda) x_j~~ \text{and} ~~{y}_{mix} = \lambda \tilde{y}_i + (1-\lambda) \tilde{y}_j,
\end{equation}
where $\lambda \in [0,1]$ is the balance parameter between two examples. Thus, \emph{mixup} extends the training distribution by updating the DNN for the constructed mini-batch.

\smallskip
\noindent {\underline{Remark}: The implicit regularization improves the generalization capability of the DNN without reducing the representational capacity. It also does not introduce sensitive model-dependent hyperparameters because it is applied to the training data. However, the extended feature or label space slows down the convergence of training.}  %However, the performance gain is relatively insignificant compared to the explicit regularization.} 

\subsection{Robust Loss Function}
\label{sec:robust_loss_function}

{
It was proven that a learned DNN with a \emph{suitably modified} loss function $\ell^{\prime}$ for noisy data $\tilde{\mathcal{D}}$ can approach the Bayes optimal classifier $f^{*}$, which achieves the optimal Bayes risk $\mathcal{R}^{*} = \mathcal{R}_{\mathcal{D}}(f^{*})$ for clean data $\mathcal{D}$. Let $\hat{f} = \text{argmin}_{f \in \mathcal{F}} \hat{\mathcal{R}}_{\ell^{\prime}, \tilde{\mathcal{D}}}(f)$ be the learned classifier with the modified loss $\ell^{\prime}$ for the noisy data, where $\hat{\mathcal{R}}_{\ell^{\prime}, \tilde{\mathcal{D}}}(f) = \mathbb{E}_{\tilde{\mathcal{D}}}[\ell(f(x;\Theta), \tilde{y})]$. If $\ell$ is $L$-Lipschitz and classification-calibrated \cite{van2015learning}, with probability at least $1\!-\!\delta$, there exists a non-decreasing function $\zeta_{\ell}$ with $\zeta_{\ell}(0)=0$ \cite{natarajan2013learning} such that \looseness=-1

\vspace*{-0.6cm}
\begin{equation}
% /zeta
\begin{split}
\mathcal{R}_{\mathcal{D}}(\hat{f}) - \mathcal{R}^{*} \leq &~\overbrace{\zeta_{\ell} \Big( \text{min}_{f\in \mathcal{F}}\mathcal{R}_{\ell,\mathcal{D}}(f) - \text{min}_{f}\mathcal{R}_{\ell, \mathcal{D}}(f)}^{\text{Approximation and Estimation Errors}}\\
&~~+ 4L_{p}\text{RC}(\mathcal{F}) + 2\sqrt{{\text{log}(1/\delta)}\big/{2|\mathcal{D}|}} \Big),\!\!\!\!\!
\end{split}
\end{equation}
% Rademacher complexity: \cite{shalev2014understanding}
$L_{p}$ is the Lipschitz constant of $\ell^{\prime}$ and RC is the Rademacher complexity of the hypothesis class $\mathcal{F}$. Then, by the universal approximation theorem \cite{csaji2001approximation}, the Bayes optimal classifier $f^*$ is guaranteed to be in the hypothesis class $\mathcal{F}$ with DNNs. %Thus, robust loss functions for DNNs have been designed using this theoretical foundation \cite{ghosh2017robust, zhang2018generalized, wang2019symmetric, lyu2020curriculum}. 

%Considering the robustness of risk minimization schemes on the loss function, 
Based on this theoretical foundation, researchers have attempted to design robust loss functions such that they achieve a small risk for unseen clean data even when noisy labels exist in the training data \cite{ghosh2017robust, zhang2018generalized, wang2019symmetric, lyu2020curriculum, feng2020can, liu2020peer}. 
}

\smallskip
\noindent \underline{Technical Detail}: % a function that satisfies the condition
Initially, Manwani and Sastry \cite{manwani2013noise} theoretically proved a sufficient condition for the loss function such that risk minimization with that function becomes noise-tolerant for binary classification. Subsequently, the sufficient condition was extended for multi-class classification using deep learning \cite{ghosh2017robust}. Specifically, a loss function is defined to be \emph{noise-tolerant} for a $c$-class classification under \emph{symmetric} noise if the function satisfies the noise rate $\tau<\frac{c-1}{c}$ and
\begin{equation}
\label{eq:symmetric_function}
\sum_{j=1}^{c}\ell\big(f(x;\Theta),y=j\big)=C, ~\forall x\in\mathcal{X}, ~\forall f,
\end{equation}
where $C$ is a constant. This condition guarantees that the classifier trained on noisy data has the same misclassification probability as that trained on noise-free data under the specified assumption. {An extension for \emph{multi-label} classification was provided by Kumar et al. \cite{kumar2020robust}.} Moreover, if $\mathcal{R}_{\mathcal{D}}(f^{*})=0$, then the function is also noise-tolerant under an \emph{asymmetric} noise, where $f^{*}$ is a global risk minimizer of $\mathcal{R}_{\mathcal{D}}$. \looseness=-1

For the classification task, the categorical cross entropy\,(CCE) loss is the most widely used loss function owing to its fast convergence and high generalization capability. However, in the presence of noisy labels, the \emph{robust MAE} \cite{ghosh2017robust} showed that the mean absolute error\,(MAE) loss achieves better generalization than the CCE loss because only the MAE loss satisfies the aforementioned condition. A limitation of the MAE loss is that its generalization performance degrades significantly when complicated data are involved. Hence, the \emph{generalized cross entropy}\,(GCE) \cite{zhang2018generalized} was proposed to achieve the advantages of both MAE and CCE losses; the GCE loss is a more general class of noise-robust loss that encompasses both of them. 
{
Amid et al. \cite{amid2019two} extended the GCE loss by introducing two temperatures based on the Tsallis divergence. \emph{Bi-tempered loss} \cite{amidrobust} introduces a proper unbiased generalization of the CE loss based on the Bregman divergence.}
In addition, inspired by the symmetricity of the Kullback-Leibler divergence, the symmetric cross entropy\,(SCE) \cite{wang2019symmetric} was proposed by combining a noise tolerance term, namely reverse cross entropy loss, with the standard CCE loss. 

Meanwhile, the \emph{curriculum loss}\,(CL) \cite{lyu2020curriculum} is a surrogate loss of the 0-1 loss function; it provides a tight upper bound and can easily be extended to multi-class classification.  {The \emph{active passive loss}\,(APL) \cite{ma2020normalized} is a combination of two types of robust loss functions, an active loss that maximizes the probability of belonging to the given class and a passive loss that minimizes the probability of belonging to other classes.}

\smallskip
\noindent \underline{Remark}: The robustness of these methods is theoretically supported well. However, they perform well only in simple cases, when learning is easy or the number of classes is small \cite{ren2018learning}. Moreover, the modification of the loss function increases the training time for convergence \cite{zhang2018generalized}.
%are significantly affected by noisy labels. Such implementations

%\begin{figure*}[ht!]
%\begin{center}
%\includegraphics[width=15.8cm]{figures/methodology/loss_adjustment_and_sample_selection_ver2.pdf}
%\end{center}
%\vspace*{-0.1cm}
%\hspace*{3.6cm} \small{(a) Loss Adjustment.} \hspace*{5.25cm} \small{(b) Sample Selection.} 
%\vspace*{-0.15cm}
%\caption{Comparison of two different training procedures: (a) shows the training procedures of %\emph{loss adjustment}; (b) shows the training procedures of \emph{sample selection}. (This figure is adapted from Song et al. \cite{song2019selfie}.)}
%\label{fig:loss_adjustment_and_sample_selection}
%\vspace*{-0.4cm}
%\end{figure*}

\vspace*{-0.2cm}
\subsection{Loss Adjustment}
\label{sec:loss_adjustment}
\vspace*{-0.0cm}

Loss adjustment is effective for reducing the negative impact of noisy labels by adjusting the loss of all training examples before updating the DNN \cite{patrini2017making, hendrycks2018using, wang2017multiclass, chang2017active, reed2015training, arazo2019unsupervised, ma2018dimensionality, song2019selfie}. The methods associated with it can be categorized into three groups depending on their adjustment philosophy: \emph{1)} {\emph{loss correction}} that estimates the noise transition matrix to correct the forward or backward loss, \emph{2)} {\emph{loss reweighting}} that imposes different importance to each example for a weighted training scheme, \emph{3)} {\emph{label refurbishment}} that adjusts the loss using the refurbished label obtained from a convex combination of noisy and predicted labels, and \emph{4)} {\emph{meta learning}} that automatically infers the optimal rule for loss adjustment. {Unlike the robust loss function newly designed for robustness, this family of methods aims to make the traditional optimization process robust to label noise. Hence, in the middle of training, the {update rule} is adjusted such that the negative impact of label noise is minimized.}

In general, loss adjustment allows for a \emph{full exploration} of the training data by adjusting the loss of every example. However, the error incurred by \emph{false} correction is accumulated, especially when the number of classes or the number of mislabeled examples is large \cite{han2018co}. 

\vspace*{0.15cm}
\subsubsection{\textbf{Loss Correction}}
\label{sec:loss_correction}
Similar to the noise adaptation layer presented in Section \ref{sec:robust_architecture}, this approach modifies the loss of each example by multiplying the estimated label transition probability by the output of a specified DNN. The main difference is that the learning of the transition probability is decoupled from that of the model. 

\smallskip
\noindent \underline{Technical Detail}:
\emph{Backward correction} \cite{patrini2017making} initially approximates the noise transition matrix using the softmax output of the DNN trained without loss correction. Subsequently, it retrains the DNN while correcting the original loss based on the estimated matrix. The corrected loss of a example $(x,\tilde{y})$ is computed by a linear combination of its loss values for observable labels, whose coefficient is the inverse transition matrix $\text{T}^{-1}$ to the observable label $y\in\{1,\dots, c\}$, given its target label $\tilde{y}$.
Therefore, the backward correction $\cev{\ell}$ is performed by multiplying the inverse transition matrix to the prediction for all the observable labels, 
{
\begin{equation}
\small
\label{eq:backward_correction}
\begin{split}
%\cev{\ell}\big(f(x;&\Theta),\tilde{y}\big) = \!\!\!\!\!\!\sum_{y \in \{1, 2, \dots, c\}}\!\!\!\!\!\!\hat{p}(\tilde{y}| y )\ell\big(f(x;\Theta),y\big)\\&=\hat{\text{T}}_{\cdot j}^{-1}\Big(\ell\big(f(x;\Theta), 1\big), \dots, \ell\big(f(x;\Theta), c\big)\Big)^{\top}\!\!,
\cev{\ell}\big(f(x;&\Theta),\tilde{y}\big) = \hat{\text{T}}^{-1}\Big\langle\ell\big(f(x;\Theta), 1\big), \dots, \ell\big(f(x;\Theta), c\big)\Big\rangle^{\!\top}\!\!,
\end{split}
\end{equation}
}
\noindent where $\hat{\text{T}}$ is the estimated noise transition matrix. 

Conversely, \emph{forward correction} \cite{patrini2017making} uses a linear combination of a DNN's softmax outputs before applying the loss function. Hence, the forward correction $\vec{\ell}$ is performed by multiplying the estimated transition probability with the softmax outputs during the forward propagation step,
{
\begin{equation}
\small
\label{eq:forward_correction}
\begin{split}
\vec{\ell}\big(f(x;\Theta),\tilde{y}\big)& = \ell\Big(\Big\langle\hat{p}(\tilde{y}|1),\dots,\hat{p}(\tilde{y}|c)\Big\rangle f(x;\Theta)^{\top},\tilde{y}\Big)\\
%&=\Big(\ell\big(\hat{\text{T}}^{\top}\!f(x;\Theta), 1) \big), \dots, \ell\big(\hat{\text{T}}^{\top}\!f(x;\Theta), c) \big), \Big).
&=\ell\big(\hat{\text{T}}^{\top}f(x;\Theta)^{\top},\tilde{y}\big).
\end{split}
\end{equation}
}

\vspace*{-0.2cm}
Furthermore, \emph{gold loss correction} \cite{hendrycks2018using} assumes the availability of clean validation data or anchor points for loss correction. Thus, a more accurate transition matrix is obtained by using them as additional information, which further improves the robustness of the loss correction. {Recently, \emph{T-Revision} \cite{xia2019anchor} provides a solution that can infer the transition matrix without anchor points, and \emph{Dual T} \cite{yao2020dual} factorizes the matrix into the product of two easy-to-estimate matrices to avoid directly estimating the noisy class posterior.}
{
Beyond the instance-independent noise assumption, Zhang et al. \cite{zhang2021approximating} introduced the instance-confidence embedding to model instance-dependent noise in estimating the transition matrix. On the other hand, Yang et al. \cite{yang2021estimating} proposed to use the Bayes optimal transition matrix estimated from the distilled examples for the instance-dependent noise transition matrix.}

\smallskip
\noindent {\underline{Remark}: The robustness of these approaches is highly dependent on how precisely the transition matrix is estimated. To acquire such a transition matrix, they require prior knowledge in general, such as anchor points or clean validation data.}

\vspace*{0.15cm}
\subsubsection{\textbf{Loss Reweighting}} Inspired by the concept of importance reweighting \cite{liu2015classification}, loss reweighting aims to assign smaller weights to the examples with false labels and greater weights to those with true labels. Accordingly, the reweighted loss on the mini-batch $\mathcal{B}_t$ is used to update the DNN, 
{
\begin{equation}
\label{eq:loss_reweighting}
\Theta_{t+1} = \Theta_{t} - \eta\nabla \Big( \frac{1}{|\mathcal{B}_{t}|}\!\sum_{(x,\tilde{y})\in\mathcal{B}_t}\!\!\!\!\overbrace{w(x,\tilde{y})\ell\big(f(x;\Theta_t), \tilde{y}\big)}^{\text{Reweighted Loss}}\Big),
\end{equation}}
where $w(x,\tilde{y})$ is the weight of an example $x$ with its noisy label $\tilde{y}$. Hence, the examples with smaller weights do not significantly affect the DNN learning.

\smallskip
\noindent \underline{Technical Detail}:
In \emph{importance reweighting} \cite{wang2017multiclass}, the ratio of two joint data distributions {$w(x,\tilde{y})=P_{\mathcal{D}}(x,\tilde{y})/P_{\tilde{\mathcal{D}}}(x,\tilde{y})$} determines the contribution of the loss of each noisy example. An approximate solution to estimate the ratio was developed because the two distributions are difficult to determine from noisy data. Meanwhile, \emph{active bias} \cite{chang2017active} emphasizes uncertain examples with inconsistent label predictions by assigning their prediction variances as the weights for training. 
{
\emph{DualGraph} \cite{zhang2021dualgraph} employs graph neural networks and reweights the examples according to the structural relations among labels, eliminating the abnormal noise examples.}

\smallskip
\noindent {\underline{Remark}: These approaches need to manually pre-specify the weighting function as well as there additional hyper-parameters, which is fairly hard to be applied in practice due to the significant variation of appropriate weighting schemes that rely on the noise type and training data.}

\vspace*{0.15cm}
\subsubsection{\textbf{Label Refurbishment}} Refurbishing a noisy label $\tilde{y}$ effectively prevents overfitting to false labels. Let $\hat{y}$ be the current prediction of the DNN $f(x;\Theta)$. Therefore, the refurbished label $y^{refurb}$ can be obtained by a convex combination of the noisy label $\tilde{y}$ and the DNN prediction $\hat{y}$,
\begin{equation}
\label{eq:label_correction}
y^{refurb} = \alpha \tilde{y} + (1-\alpha) \hat{y},
\end{equation}
where $\alpha \in [0,1]$ is the label confidence of $\tilde{y}$. To mitigate the damage of incorrect labeling, this approach backpropagates the loss for the refurbished label instead of the noisy one, thereby yielding substantial robustness to noisy labels. 

\smallskip
\noindent \underline{Technical Detail}:
\emph{Bootstrapping} \cite{reed2015training} is the first method that proposes the concept of label refurbishment to update the target label of training examples. It develops a more coherent network that improves its ability to evaluate the consistency of noisy labels, with the label confidence $\alpha$ obtained via cross-validation.
\emph{Dynamic bootstrapping} \cite{arazo2019unsupervised} dynamically adjusts the confidence $\alpha$ of individual training examples. The confidence $\alpha$ is obtained by fitting a two-component and one-dimensional beta mixture model to the loss distribution of all training examples. {\emph{Self-adaptive training} \cite{huang2020self} applies the exponential moving average to alleviate the instability issue of using instantaneous prediction of the current DNN, % scheme
\begin{equation}
\!\!y_{t+1}^{refurb} = \alpha y_t^{refurb} + (1-\alpha)\hat{y}, ~\text{where}~ y_0^{refurb} = \tilde{y}\!\!
\end{equation}
}
% instantaneous
%For each training epoch, it estimates the probability of an example $x$ being true-labeled by fitting a two-component and one-dimensional beta mixture model to the loss distribution of all training examples and then uses it as the confidence $\alpha$. As true-labeled examples exhibit smaller losses than false-labeled ones, {\color{blue} [Need to change] the confidence $\alpha$ of an example $x$ is obtained through the posterior probability $p(g|\ell(f(x;\Theta_t), \tilde{y}))$ of the mixture model, where $g$ is the beta component with a smaller mean.}
\emph{D2L} \cite{ma2018dimensionality} trains a DNN using a dimensionality-driven learning strategy to avoid overfitting to false labels. A simple measure called \emph{local intrinsic dimensionality} \cite{houle2017local} is adopted to evaluate the confidence $\alpha$ in considering that the overfitting is exacerbated by dimensional expansion. Hence, refurbished labels are generated to prevent the dimensionality of the representation subspace from expanding at a later stage of training. 
Recently, \emph{SELFIE} \cite{song2019selfie} introduces a novel concept of \emph{refurbishable examples} that can be corrected with high precision. The key idea is to consider the example with consistent label predictions as refurbishable because such consistent predictions correspond to its true label with a high probability owing to the learner's perceptual consistency. Accordingly, the labels of only refurbishable examples are corrected to minimize the number of falsely corrected cases. {Similarly, \emph{AdaCorr} \cite{zheng2020error} selectively refurbishes the label of noisy examples, but a theoretical error-bound is provided. Alternatively, \emph{SEAL} \cite{chen2021beyond} averages the softmax output of a DNN on each example over the whole training process, then re-trains the DNN using the averaged soft labels.}

\smallskip
\noindent {\underline{Remark}: Differently from loss correction and reweighting, all the noisy labels are explicitly replaced with other expected clean labels (or their combination). If there are not many confusing classes in data, these methods work well by refurbishing the noisy labels with high precision. In the opposite case, the DNN could overfit to wrongly refurbished labels.}

\vspace*{0.15cm}
\subsubsection{\textbf{Meta Learning}}
{In recent years, meta learning becomes an important topic in the machine learning community and is applied to improve noise robustness \cite{finn2017model, shu2019meta, wang2020training}. The key concept is \emph{learning to learn} that performs learning at a level higher than conventional learning, thus achieving data-agnostic and noise type-agnostic rules for better practical use. It is similar to loss reweighting and label refurbishment, but the adjustment is automated in a meta learning manner.

\smallskip
\noindent \underline{Technical Detail}:
For the loss reweighting in Eq.~\eqref{eq:loss_reweighting}, the goal is to learn the weight function $w(x,\tilde{y})$. Specifically, \emph{L2LWS} \cite{dehghani2017learning} and \emph{CWS} \cite{dehghani2017avoiding} are unified neural architectures composed of a target DNN and a meta-DNN. The meta-DNN is trained on a small clean validation dataset; it then provides guidance to evaluate the weight score for the target DNN. Here, part of the two DNNs are shared and jointly trained to benefit from each other. \emph{Automatic reweighting} \cite{ren2018learning} is a meta learning algorithm that learns the weights of training examples based on their gradient directions. It includes a small clean validation dataset into the training dataset and reweights the backward loss of the mini-batch examples such that the updated gradient minimizes the loss of this validation dataset.  \emph{Meta-weight-net} \cite{shu2019meta} parameterizes the weighting function as a multi-layer perceptron network with only one hidden layer. A meta-objective is defined to update its parameters such that they minimize the empirical risk of a small clean dataset. At each iteration, the parameter of the target network is guided by the weight function updated via the meta-objective. 
{Likewise, \emph{data coefficients}\,(i.e., exemplar weights and true labels) \cite{zhang2020distilling} are estimated by meta-optimization with a small clean set, which is only $0.2$\% of the entire training set, while refurbishing the examples probably mislabeled.}

For the label refurbishment in Eq.~\eqref{eq:label_correction}, \emph{knowledge distillation} \cite{li2017learning} adopts the technique of transferring knowledge from one expert model to a target model. The prediction from the expert DNN trained on small clean validation data is used instead of the prediction $\hat{y}$ from the target DNN. %In addition, it leverages a knowledge graph, which encodes the structure of the label space, to further elaborate the expert DNN's prediction.
{\emph{MLC} \cite{zheng2021meta} updates the target model with corrected labels provided by a meta model trained on clean validation data. The two models are trained concurrently via a bi-level optimization.} 

\smallskip
\noindent \underline{Remark}:
By learning the update rule via meta learning, the trained network easily adapts to various types of data and label noise. Nevertheless, unbiased clean validation data is essential to minimize the auxiliary objective, although it may not be available in real-world data.}

% 1. co-training, 2. iterative training, 3. extension to semi-supervised learning.
\vspace*{-0.1cm}
\subsection{Sample Selection} 
\label{sec:sample_selection}

To avoid any false corrections, many recent studies \cite{malach2017decoupling, jiang2018mentornet, han2018co, yu2019does, wang2018iterative, shen2019learning, chen2019understanding, song2019selfie, nguyen2020self, lyu2020curriculum, song2020two} have adopted sample selection that involves selecting true-labeled examples from a noisy training dataset. In this case, the update equation in Eq.\ \eqref{eq:corrupted_update} is modified to render a DNN more robust for noisy labels. Let $\mathcal{C}_t \subseteq \mathcal{B}_t$ be the identified \emph{clean} examples at time $t$. Then, the DNN is updated only for the selected clean examples $\mathcal{C}_t$, 
\begin{equation}
\label{eq:clean_update}
\Theta_{t+1} = \Theta_{t} - \eta\nabla\Big(\frac{1}{|\mathcal{C}_t|} \!\sum_{(x,\tilde{y}) \in \mathcal{C}_t} \!\!\!\!\ell\big(f(x;\Theta_{t}), \tilde{y}\big)\Big),
\end{equation}
where the rest mini-batch examples, which are likely to be false-labeled, are excluded to pursue robust learning. 

\newcommand{\norm}[1]{\left\lVert#1\right\rVert}
{The {memorization nature} of DNNs has been explored theoretically and empirically to identify clean examples from noisy training data \cite{krueger2017deep, zhang2019identity, yao2020searching}.
Specifically, assuming clusterable data where the clusters are located on the unit Euclidean ball, Li et al. \cite{li2020gradient} proved the distance from the initial weight ${W}_{0}$ to the weight ${W}_t$ after $t$ iterations,
\begin{equation}
\label{eq:memorization_effect_foundation}
\norm{{W}_t - {W}_0}_{F} \lesssim \big( \sqrt{K} + (K^{2}\epsilon_{0}/\norm{{C}}^{2})t \big),
\end{equation}
where $\norm{\cdot}_{F}$ is the Frobenius norm, $K$ is the number of clusters, and ${C}$ is the set of cluster centers reaching all input examples within their $\epsilon_0$ neighborhood.
Eq.\ \eqref{eq:memorization_effect_foundation} demonstrates that the weights of DNNs start to stray far from the initial weights when overfitting to corrupted labels, while they are still in the vicinity of the initial weights at an early stage of training \cite{li2020gradient, han2020survey}. In the empirical studies \cite{arpit2017closer, song2019prestopping}, the \emph{memorization effect} is also observed since DNNs tend to first learn simple and generalized patterns and then gradually overfit to all noisy patterns. As such, favoring small-loss training examples as the clean ones are commonly employed to design robust training methods \cite{han2018co, jiang2018mentornet, chen2019understanding, shen2019learning, li2020dividemix}.

Learning with sample selection is well motivated and works well in general, but this approach suffers from accumulated error caused by incorrect selection, especially when there are many ambiguous classes in training data.} 
Hence, recent approaches often leverage multiple DNNs to cooperate with one another \cite{han2018co} or run multiple training rounds \cite{wang2018iterative}.
Moreover, to benefit from even false-labeled examples, loss correction or semi-supervised learning have been recently combined with the sample selection strategy \cite{song2019selfie, li2020dividemix}.

\begin{figure}
\vspace*{+0.08cm}
\begin{center}
\includegraphics[width=8.7cm]{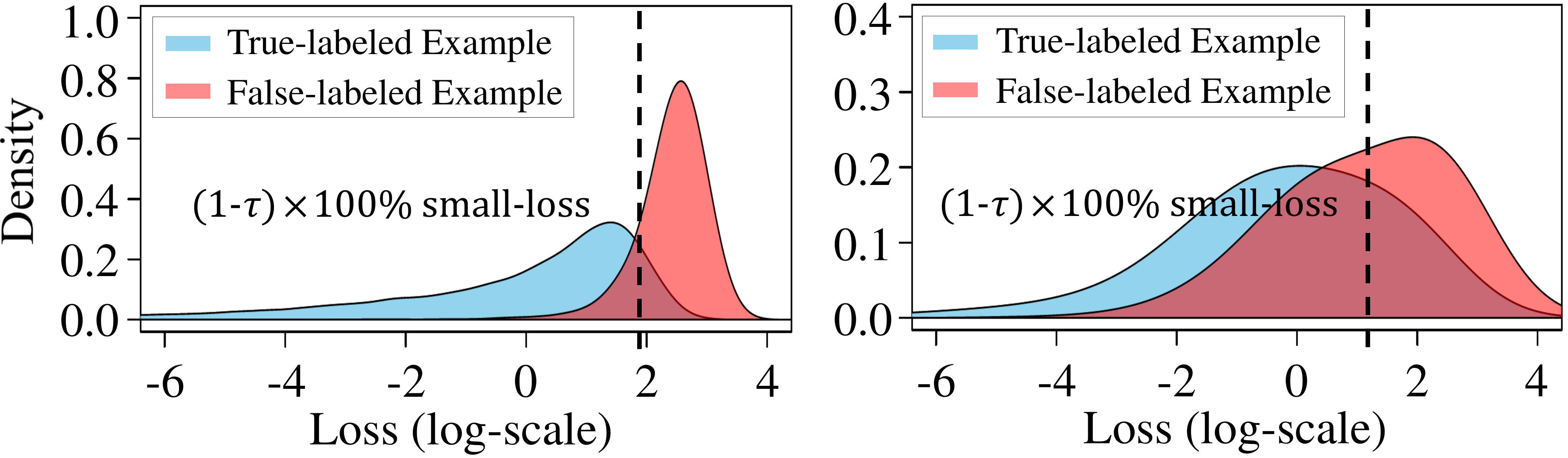}
\end{center}
\vspace*{-0.2cm}
\hspace*{0.80cm} {\small (a) Symmetric Noise $40\%$.} \hspace*{0.525cm} {\small (b) Asymmetric Noise $40\%$.}
\vspace*{-0.15cm}
\caption{Loss distribution of training examples at the training accuracy of $50\%$ on noisy CIFAR-100. (This figure is adapted from Song et al. \cite{song2019prestopping}.)}
\label{fig:loss_distribution}
\vspace*{-0.4cm}
\end{figure}

\smallskip\smallskip
\subsubsection{\textbf{Multi-network Learning}}
{
Collaborative learning and co-training are widely used for the multi-network training. Consequently, the sample selection process is guided by the mentor network in the case of collaborative learning or the peer network in the case of co-training.
}

\smallskip
\noindent \underline{Technical Detail}:
Initially, \emph{Decouple} \cite{malach2017decoupling} proposes the decoupling of when to update from how to update. Hence, two DNNs are maintained simultaneously and updated only the examples selected based on a disagreement between the two DNNs.
Next, due to the memorization effect of DNNs, many researchers have adopted another selection criterion, called a \emph{small-loss} trick, which treats a certain number of small-loss training examples as true-labeled examples; many true-labeled examples tend to exhibit smaller losses than false-labeled examples, as illustrated in Figure \ref{fig:loss_distribution}(a). In \emph{MentorNet} \cite{jiang2018mentornet}, a pre-trained mentor network guides the training of a student network in a collaborative learning manner. Based on the small-loss trick, the mentor network provides the student network with examples whose labels are likely to be correct. \emph{Co-teaching} \cite{han2018co} and \emph{Co-teaching+} \cite{yu2019does} also maintain two DNNs, but each DNN selects a certain number of small-loss examples and feeds them to its peer DNN for further training. \emph{Co-teaching+} further employs the disagreement strategy of \emph{Decouple} compared with \emph{Co-teaching}. {In contrast, \emph{JoCoR} \cite{wei2020combating} reduces the diversity of two networks via co-regularization, making predictions of the two networks closer.}

\smallskip
\noindent {\underline{Remark}: The co-training methods help reduce the confirmation bias \cite{han2018co}, which is a hazard of favoring the examples selected at the beginning of training, while the increase in the number of learnable parameters makes their learning pipeline inefficient. In addition, the small-loss trick does not work well when the loss distribution of true-labeled and false-labeled examples largely overlap, as in the asymmetric noise in Figure \ref{fig:loss_distribution}(b).}

\smallskip\smallskip
\subsubsection{\textbf{Multi-round Learning}}
Without maintaining additional DNNs, multi-round learning iteratively refines the selected set of clean examples by repeating the training round. Thus, the selected set keeps improved as the number of rounds increases.

\smallskip
\noindent \underline{Technical Detail}:
\emph{ITLM} \cite{shen2019learning} iteratively minimizes the trimmed loss by alternating between selecting true-labeled examples at the current moment and retraining the DNN using them. At each training round, only a fraction of small-loss examples obtained in the current round are used to retrain the DNN in the next round. \emph{INCV} \cite{chen2019understanding} randomly divides noisy training data and then employs cross-validation to classify true-labeled examples while removing large-loss examples at each training round. Here, \emph{Co-teaching} is adopted to train the DNN on the identified examples in the final round of training. {Similarly, \emph{O2U-Net} \cite{huang2019o2u} repeats the whole training process with the cyclical learning rate until enough loss statistics of every examples are gathered. Next, the DNN is re-trained from scratch only for the clean data where false-labeled examples have been detected and removed based on statistics.} \looseness=-1

A number of variations have been proposed to achieve high performance using iterative refinement only in a single training round.
Beyond the small-loss trick, \emph{iterative detection} \cite{wang2018iterative} detects false-labeled examples by employing the local outlier factor algorithm \cite{breunig2000lof}. With a Siamese network, it gradually pulls away false-labeled examples from true-labeled samples in the deep feature space. \emph{MORPH} \cite{song2020two} introduces the concept of memorized examples which is used to iteratively expand an initial safe set into a maximal safe set via self-transitional learning. \emph{TopoFilter} \cite{wu2020topological} utilizes the spatial topological pattern of learned representations to detect true-labeled examples, not relying on the prediction of the noisy classifier.
{
\emph{NGC} \cite{wu2021ngc} iteratively constructs the nearest neighbor graph using latent representations and performs geometry-based sample selection by aggregating information from neighborhoods. Soft pesudo-labels are assigned to the examples not selected.
}

\smallskip
\noindent {\underline{Remark}: The selected clean set keeps expanded and purified with iterative refinement, mainly through multi-round learning. As a side effect, the computational cost for training increases linearly for the number of training rounds.}

\smallskip
\subsubsection{\textbf{Hybrid Approach}} {An inherent limitation of sample selection is to discard all the \emph{unselected} training examples, thus resulting in a \emph{partial} exploration of training data. To exploit all the noisy examples, researchers have attempted to combine sample selection with other orthogonal ideas.

\begin{figure}[t!]
\begin{center}
\includegraphics[width=8.45cm]{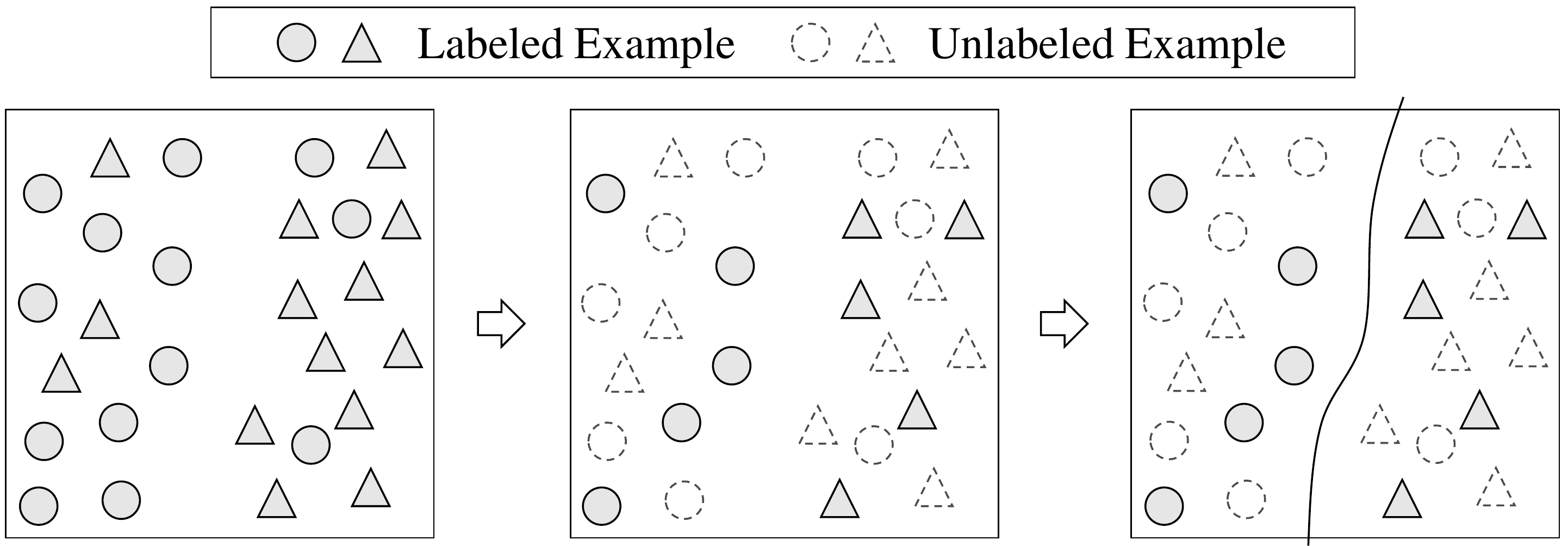}
\end{center}
\vspace*{-0.1cm}
\hspace*{0.2cm} \small{(a) Noisy Data.} \hspace*{0.4cm} \small{(b) Transformed Data.} \hspace*{0.8cm} \small{(c) SSL.}
\vspace*{-0.2cm}
\caption{{Procedures for semi-supervised learning under label noise.}}
\label{fig:noise_semi_supervised}
\vspace*{-0.45cm}
\end{figure}

\smallskip
\noindent \underline{Technical Detail}:
The most prominent method in this direction is combining a specific sample selection strategy with a specific semi-supervised learning model. As illustrated in Figure \ref{fig:noise_semi_supervised}, selected examples are treated as labeled clean data, whereas the remaining examples are treated as unlabeled. Subsequently, semi-supervised learning is performed using the transformed data. \emph{SELF} \cite{nguyen2020self} is combined with a semi-supervised learning approach to progressively filter out false-labeled examples from noisy data. By maintaining the running average model called the {mean-teacher} \cite{tarvainen2017mean} as the backbone, it obtains the self-ensemble predictions of all training examples and then progressively removes examples whose ensemble predictions do not agree with their annotated labels. This method further leverages unsupervised loss from the examples not included in the selected clean set. \emph{DivideMix} \cite{li2020dividemix} uses two-component and one-dimensional Gaussian mixture models to transform noisy data into labeled (clean) and unlabeled (noisy) sets. Then, it applies a semi-supervised technique \emph{MixMatch} \cite{berthelot2019mixmatch}. Recently, \emph{RoCL} \cite{zhou2021robust} employs two-phase learning strategies: supervised training on selected clean examples and then semi-supervised learning on relabeled noisy examples with self-supervision. For selection and relabeling, it computes the exponential moving average of the loss over training iterations. 
}

\clearpage
{

\newcolumntype{L}[1]{>{\centering\let\newline\\\arraybackslash\hspace{0pt}}m{#1}}
\newcolumntype{X}[1]{>{\centering\let\newline\\\arraybackslash\hspace{0pt}}p{#1}}
\newcolumntype{Y}[1]{>{\let\newline\\\arraybackslash\hspace{1pt}}m{#1}}

\begin{table*}[h]
\vspace*{-3.9cm}
\caption{{Comparison of proposed robust deep learning methods with respect to the following six properties: (P1)\,Flexibility, (P2)\,No Pre-training, (P3)\,Full Exploration, (P4)\,No Supervision, (P5)\,Heavy Noise, and (P6)\,Complex Noise.}}
\vspace*{-0.2cm}
\begin{tabular}{L{0.7cm}|L{2.5cm}|Y{4.7cm}|X{0.5cm}|X{0.5cm}|X{0.5cm}|X{0.5cm}|X{0.5cm}|X{0.5cm}|Y{2.9cm}}
\toprule

\multicolumn{2}{c|}{\textbf{Category}} & \hspace*{2.0cm}\textbf{Method} & \textbf{P1} & \textbf{P2} & \textbf{P3} & \textbf{P4} & \textbf{P5} & \textbf{P6} & \,\,\,\,\,\,\,\,\,\textbf{Implementation}\!\! \\ \hline

\multirow{9}{*}{\vspace*{-0.35cm}\hspace*{0.13cm}\!\!\!\!\!\!\rotatebox[origin=c]{90}{\makecell[l]{Robust Architecture\\\hspace*{0.66cm}(\textsection \ref{sec:robust_architecture})}}\!\!\!\!\!} 
& \multirow{6}{*}{\vspace*{-0.2cm}\makecell[l]{Noisy Adaptation\\\hspace*{0.67cm}Layer}}  
& \emph{Webly Learning}\cite{chen2015webly} & \cellcolor{gray!10}$\bigtriangleup$ & \cellcolor{red!10}\xmark & \cellcolor{blue!10}\cmark & \cellcolor{blue!10}\cmark & \cellcolor{red!10}\xmark & \cellcolor{red!10}\xmark & Official\,(Caffe)\protect\footnotemark[1] \\\cline{3-10} % 
& & \emph{Noise Model}\cite{sukhbaatar2014training} & \cellcolor{gray!10}$ \bigtriangleup$ & \cellcolor{blue!10}\cmark & \cellcolor{blue!10}\cmark & \cellcolor{blue!10}\cmark & \cellcolor{red!10}\xmark & \cellcolor{red!10}\xmark & Unofficial\,(Keras)\protect\footnotemark[2] \\\cline{3-10}
& & \emph{Dropout Noise Model}\cite{jindal2016learning} & \cellcolor{gray!10}$ \bigtriangleup$ & \cellcolor{blue!10}\cmark & \cellcolor{blue!10}\cmark & \cellcolor{blue!10}\cmark & \cellcolor{red!10}\xmark & \cellcolor{red!10}\xmark & Official\,(MATLAB)\protect\footnotemark[3] \\\cline{3-10}
& & \emph{S-model}\cite{goldberger2017training} & \cellcolor{gray!10}$ \bigtriangleup$ & \cellcolor{blue!10}\cmark & \cellcolor{blue!10}\cmark & \cellcolor{blue!10}\cmark & \cellcolor{red!10}\xmark & \cellcolor{red!10}\xmark & Official\,(Keras)\protect\footnotemark[4]\\\cline{3-10}
& & \emph{C-model}\cite{goldberger2017training} & \cellcolor{gray!10}$ \bigtriangleup$ & \cellcolor{blue!10}\cmark & \cellcolor{blue!10}\cmark & \cellcolor{blue!10}\cmark & \cellcolor{red!10}\xmark & \cellcolor{blue!10}\cmark & Official\,(Keras)\protect\footnotemark[4] \\\cline{3-10}
& & \emph{NLNN}\cite{bekker2016training} & \cellcolor{gray!10}$ \bigtriangleup$ & \cellcolor{blue!10}\cmark & \cellcolor{blue!10}\cmark & \cellcolor{blue!10}\cmark & \cellcolor{red!10}\xmark & \cellcolor{red!10}\xmark & Unofficial\,(Chainer)\protect\footnotemark[5] \\\cmidrule{2-10}

& \multirow{4}{*}{\makecell[l]{\hspace*{0.12cm}Dedicated\\Architecture}} 
& \emph{Probablistic Noise Model}\cite{xiao2015learning} & \cellcolor{red!10}\xmark & \cellcolor{red!10}\xmark & \cellcolor{blue!10}\cmark & \cellcolor{red!10}\xmark & $ \cellcolor{gray!10}\bigtriangleup$ & \cellcolor{blue!10}\cmark & Official\,(Caffe)\protect\footnotemark[6] \\\cline{3-10}
& & \emph{Masking}\cite{han2018masking} & \cellcolor{red!10}\xmark & \cellcolor{blue!10}\cmark & \cellcolor{blue!10}\cmark & \cellcolor{red!10}\xmark & $ \cellcolor{gray!10}\bigtriangleup$ & \cellcolor{blue!10}\cmark & Official\,(TensorFlow)\protect\footnotemark[7] \\\cline{3-10}
& & \emph{Contrastive-Additive Noise Network}\cite{yao2018deep} & \cellcolor{red!10}\xmark & \cellcolor{blue!10}\cmark & \cellcolor{blue!10}\cmark &\cellcolor{blue!10}\cmark & $ \cellcolor{gray!10}\bigtriangleup$ &\cellcolor{blue!10}\cmark & N/A \\\cline{3-10}
& & {\emph{RoG}\cite{lee2019robust}} & \cellcolor{blue!10}\cmark & \cellcolor{red!10}\xmark & \cellcolor{blue!10}\cmark &\cellcolor{blue!10}\cmark & \cellcolor{blue!10}\cmark &$ \cellcolor{gray!10}\bigtriangleup$ & Official\,(PyTorch)\protect\footnotemark[8] \\\midrule

\multirow{9}{*}{\vspace*{-0.25cm}\hspace*{0.13cm}\!\!\!\!\!\!\rotatebox[origin=c]{90}{\makecell[l]{Robust Regularization\\\hspace*{0.8cm}(\textsection \ref{sec:robust_regularization})}}\!\!\!\!\!} 
& \multirow{6}{*}{\makecell[l]{\hspace*{0.35cm}Explicit\\Regularization}}  
& \emph{Bilevel Learning}\cite{jenni2018deep} & \cellcolor{blue!10}\cmark & \cellcolor{blue!10}\cmark & \cellcolor{blue!10}\cmark & \cellcolor{red!10}\xmark & \cellcolor{gray!10}$ \bigtriangleup$ & \cellcolor{gray!10}$ \bigtriangleup$ & Official\,(TensorFlow)\protect\footnotemark[9] \\\cline{3-10}
& & \emph{Annotator Confusion}\cite{tanno2019learning} & \cellcolor{blue!10}\cmark & \cellcolor{red!10}\xmark & \cellcolor{blue!10}\cmark & \cellcolor{blue!10}\cmark & \cellcolor{gray!10}$ \bigtriangleup$ & \cellcolor{gray!10}$ \bigtriangleup$ & Official\,(TensorFlow)\protect\footnotemark[10] \\\cline{3-10}
& & \emph{Pre-training}\cite{hendrycks2019using} & \cellcolor{blue!10}\cmark & \cellcolor{red!10}\xmark & \cellcolor{blue!10}\cmark & \cellcolor{blue!10}\cmark & \cellcolor{gray!10}$ \bigtriangleup$ & \cellcolor{gray!10}$ \bigtriangleup$ & Official\,(PyTorch)\protect\footnotemark[11] \\\cline{3-10}
& & {\emph{PHuber}\cite{menon2020can}} & \cellcolor{blue!10}\cmark & \cellcolor{blue!10}\cmark  & \cellcolor{blue!10}\cmark & \cellcolor{blue!10}\cmark & \cellcolor{gray!10}$ \bigtriangleup$ & \cellcolor{gray!10}$ \bigtriangleup$ & Unofficial\,(PyTorch)\protect\footnotemark[12] \\\cline{3-10}
& & {\emph{Robust Early-learning}\cite{xia2021robust}} & \cellcolor{blue!10}\cmark & \cellcolor{blue!10}\cmark & \cellcolor{blue!10}\cmark & \cellcolor{blue!10}\cmark & \cellcolor{gray!10}$ \bigtriangleup$ & \cellcolor{gray!10}$ \bigtriangleup$ & Official\,(PyTorch)\protect\footnotemark[13] \\\cline{3-10}
& & {\emph{ODLN}\cite{wei2021open}} & \cellcolor{blue!10}\cmark & \cellcolor{blue!10}\cmark & \cellcolor{blue!10}\cmark & \cellcolor{blue!10}\cmark & \cellcolor{gray!10}$ \bigtriangleup$ & \cellcolor{gray!10}$ \bigtriangleup$ & Official\,(PyTorch)\protect\footnotemark[14] \\\cmidrule{2-10}

& \multirow{3}{*}{\makecell[l]{\hspace*{0.35cm}Implicit\\Regularization}} 
& \emph{Adversarial Training}\cite{goodfellow2014explaining} & \cellcolor{blue!10}\cmark & \cellcolor{blue!10}\cmark & \cellcolor{blue!10}\cmark & \cellcolor{blue!10}\cmark & \cellcolor{red!10}\xmark  & \cellcolor{gray!10}$ \bigtriangleup$ & Unofficial\,(PyTorch)\protect\footnotemark[15] \\\cline{3-10}
& & \emph{Label Smoothing}\cite{pereyra2017regularizing} & \cellcolor{blue!10}\cmark & \cellcolor{blue!10}\cmark & \cellcolor{blue!10}\cmark & \cellcolor{blue!10}\cmark & \cellcolor{red!10}\xmark  & \cellcolor{gray!10}$ \bigtriangleup$ & Unofficial\,(PyTorch)\protect\footnotemark[16] \\\cline{3-10}
& & \emph{Mixup}\cite{zhang2018mixup} & \cellcolor{blue!10}\cmark & \cellcolor{blue!10}\cmark & \cellcolor{blue!10}\cmark & \cellcolor{blue!10}\cmark & \cellcolor{red!10}\xmark  & \cellcolor{gray!10}$ \bigtriangleup$ & Official\,(PyTorch)\protect\footnotemark[17] \\\midrule

\multicolumn{2}{c|}{\multirow{7}{*}{{\makecell[l]{Robust Loss Function\\\hspace*{0.75cm}(\textsection \ref{sec:robust_loss_function})}}}}  
& \emph{Robust MAE}\cite{ghosh2017robust} & \cellcolor{blue!10}\cmark & \cellcolor{blue!10}\cmark & \cellcolor{blue!10}\cmark & \cellcolor{blue!10}\cmark & \cellcolor{red!10}\xmark & \cellcolor{red!10}\xmark & N/A \\\cline{3-10}
\multicolumn{2}{c|}{} & \emph{Generalized Cross Entropy}\cite{zhang2018generalized} & \cellcolor{blue!10}\cmark & \cellcolor{blue!10}\cmark & \cellcolor{blue!10}\cmark & \cellcolor{blue!10}\cmark & \cellcolor{red!10}\xmark & \cellcolor{red!10}\xmark & Unofficial\,(PyTorch)\protect\footnotemark[18] \\\cline{3-10}
\multicolumn{2}{c|}{} & \emph{Symmetric Cross Entropy}\cite{wang2019symmetric} & \cellcolor{blue!10}\cmark & \cellcolor{blue!10}\cmark & \cellcolor{blue!10}\cmark & \cellcolor{blue!10}\cmark & \cellcolor{red!10}\xmark & \cellcolor{red!10}\xmark & Official\,(Keras)\protect\footnotemark[19] \\\cline{3-10}
\multicolumn{2}{c|}{} & {\emph{Bi-tempered Loss}\cite{amidrobust}} & \cellcolor{blue!10}\cmark & \cellcolor{blue!10}\cmark & \cellcolor{blue!10}\cmark & \cellcolor{blue!10}\cmark & \cellcolor{gray!10}$\bigtriangleup$  & \cellcolor{gray!10}$\bigtriangleup$ & Official\,(TensorFlow)\protect\footnotemark[20] \\\cline{3-10}
\multicolumn{2}{c|}{} & \emph{Curriculum Learning}\cite{lyu2020curriculum} & \cellcolor{blue!10}\cmark & \cellcolor{blue!10}\cmark & \cellcolor{blue!10}\cmark & \cellcolor{red!10}\xmark & \cellcolor{blue!10}\cmark & \cellcolor{gray!10}$\bigtriangleup$ & N/A \\\cline{3-10}
\multicolumn{2}{c|}{} & {\emph{Active Passive Loss}\cite{ma2020normalized}} & \cellcolor{blue!10}\cmark & \cellcolor{blue!10}\cmark & \cellcolor{blue!10}\cmark & \cellcolor{blue!10}\cmark & \cellcolor{blue!10}\cmark & \cellcolor{gray!10}$\bigtriangleup$ & Official\,(PyTorch)\protect\footnotemark[21] \\\midrule

\multirow{21}{*}{\vspace*{-0.7cm}\hspace*{0.13cm}\!\!\!\!\!\!\rotatebox[origin=c]{90}{\makecell[l]{Loss Adjustment\\\hspace*{0.5cm}(\textsection \ref{sec:loss_adjustment})}}\!\!\!\!\!} 
& \multirow{5}{*}{\makecell[l]{Loss Correction}}  
& \emph{Backward Correction}\cite{patrini2017making} & \cellcolor{blue!10}\cmark & \cellcolor{blue!10}\cmark & \cellcolor{blue!10}\cmark &  \cellcolor{red!10}\xmark &  \cellcolor{red!10}\xmark &  \cellcolor{red!10}\xmark & Official\,(Keras)\protect\footnotemark[22] \\\cline{3-10}
& & \emph{Forward Correction}\cite{patrini2017making} & \cellcolor{blue!10}\cmark & \cellcolor{blue!10}\cmark & \cellcolor{blue!10}\cmark &  \cellcolor{red!10}\xmark &  \cellcolor{red!10}\xmark &  \cellcolor{red!10}\xmark & Official\,(Keras)\protect\footnotemark[22] \\\cline{3-10}
& & \emph{Gold Loss Correction}\cite{hendrycks2018using} & \cellcolor{blue!10}\cmark & \cellcolor{red!10}\xmark & \cellcolor{blue!10}\cmark &  \cellcolor{red!10}\xmark &  \cellcolor{red!10}\xmark &  \cellcolor{red!10}\xmark & Official\,(PyTorch)\protect\footnotemark[23] \\\cline{3-10}
& & {\emph{T-revision}\cite{xia2019anchor}} & \cellcolor{blue!10}\cmark &  \cellcolor{red!10}\xmark  & \cellcolor{blue!10}\cmark &  \cellcolor{blue!10}\cmark &  \cellcolor{red!10}\xmark  &  \cellcolor{red!10}\xmark & Official\,(PyTorch)\protect\footnotemark[24] \\\cline{3-10}
& & {\emph{Dual T}\cite{yao2020dual}} & \cellcolor{blue!10}\cmark & \cellcolor{red!10}\xmark & \cellcolor{blue!10}\cmark &  \cellcolor{blue!10}\cmark &  \cellcolor{gray!10}$ \bigtriangleup$ &  \cellcolor{red!10}\xmark & N/A \\\cmidrule{2-10}

& \multirow{3}{*}{\makecell[l]{Loss Reweigting}} 
& \emph{Importance Reweighting}\cite{wang2017multiclass} & \cellcolor{blue!10}\cmark & \cellcolor{blue!10}\cmark & \cellcolor{blue!10}\cmark & \cellcolor{blue!10}\cmark & \cellcolor{red!10}\xmark & \cellcolor{gray!10}$ \bigtriangleup$ & Unofficial\,(PyTorch)\protect\footnotemark[25] \\\cline{3-10}
& & \emph{Active Bias}\cite{chang2017active} & \cellcolor{blue!10}\cmark & \cellcolor{blue!10}\cmark & \cellcolor{blue!10}\cmark & \cellcolor{blue!10}\cmark & \cellcolor{red!10}\xmark & \cellcolor{gray!10}$ \bigtriangleup$ & Unofficial\,(TensorFlow)\protect\footnotemark[26]\!\! \\\cline{3-10}
& & {\emph{DualGraph}\cite{zhang2021dualgraph}} & \cellcolor{red!10}\xmark & \cellcolor{blue!10}\cmark & \cellcolor{blue!10}\cmark & \cellcolor{blue!10}\cmark & \cellcolor{blue!10}\cmark & \cellcolor{gray!10}$ \bigtriangleup$ & N/A \\\cmidrule{2-10}

& \multirow{6}{*}{\makecell[l]{Label Refurbishment}} 
& \emph{Bootstrapping}\cite{reed2015training} & \cellcolor{blue!10}\cmark & \cellcolor{blue!10}\cmark & \cellcolor{blue!10}\cmark & \cellcolor{red!10}\xmark & \cellcolor{red!10}\xmark & \cellcolor{gray!10}$ \bigtriangleup$ & Unofficial\,(Keras)\protect\footnotemark[27] \\\cline{3-10}
& & \emph{Dynamic Bootstrapping}\cite{arazo2019unsupervised} & \cellcolor{blue!10}\cmark & \cellcolor{blue!10}\cmark & \cellcolor{blue!10}\cmark & \cellcolor{blue!10}\cmark & \cellcolor{red!10}\xmark & \cellcolor{gray!10}$ \bigtriangleup$ & Official\,(PyTorch)\protect\footnotemark[28] \\\cline{3-10}
& & {\emph{Self-adaptive Training}\cite{huang2020self}} & \cellcolor{blue!10}\cmark & \cellcolor{red!10}\xmark & \cellcolor{blue!10}\cmark & \cellcolor{blue!10}\cmark & \cellcolor{blue!10}\cmark  & \cellcolor{gray!10}$ \bigtriangleup$ & Official\,(PyTorch)\protect\footnotemark[29] \\\cline{3-10}
& & \emph{D2L}\cite{ma2018dimensionality} & \cellcolor{blue!10}\cmark & \cellcolor{blue!10}\cmark & \cellcolor{blue!10}\cmark & \cellcolor{blue!10}\cmark & \cellcolor{red!10}\xmark & \cellcolor{gray!10}$ \bigtriangleup$ & Official\,(Keras)\protect\footnotemark[30] \\\cline{3-10}
& & {\emph{AdaCorr}\cite{zheng2020error}} & \cellcolor{blue!10}\cmark & \cellcolor{blue!10}\cmark & \cellcolor{blue!10}\cmark & \cellcolor{blue!10}\cmark & \cellcolor{blue!10}\cmark & \cellcolor{gray!10}$ \bigtriangleup$ & Official\,(PyTorch)\protect\footnotemark[31] \\\cline{3-10}
& & {\emph{SEAL}\cite{chen2021beyond}} & \cellcolor{blue!10}\cmark & \cellcolor{red!10}\xmark & \cellcolor{blue!10}\cmark & \cellcolor{blue!10}\cmark & \cellcolor{blue!10}\cmark & \cellcolor{blue!10}\cmark & Official\,(PyTorch)\protect\footnotemark[32] \\\cmidrule{2-10}

& \multirow{7}{*}{\makecell[l]{Meta Learning}} 
& \emph{L2LWS}\cite{dehghani2017learning} & \cellcolor{red!10}\xmark & \cellcolor{blue!10}\cmark & \cellcolor{blue!10}\cmark &  \cellcolor{red!10}\xmark & \cellcolor{gray!10}$ \bigtriangleup$ & \cellcolor{gray!10}$ \bigtriangleup$ & Unofficial\,(TensorFlow)\protect\footnotemark[33]\!\! \\\cline{3-10}
& & \emph{CWS}\cite{dehghani2017avoiding} & \cellcolor{red!10}\xmark & \cellcolor{blue!10}\cmark & \cellcolor{blue!10}\cmark &  \cellcolor{red!10}\xmark & \cellcolor{gray!10}$ \bigtriangleup$ & \cellcolor{gray!10}$ \bigtriangleup$ & N/A \\\cline{3-10}
& & \emph{Automatic Reweighting}\cite{ren2018learning} & \cellcolor{blue!10}\cmark & \cellcolor{blue!10}\cmark & \cellcolor{blue!10}\cmark &  \cellcolor{red!10}\xmark & \cellcolor{gray!10}$ \bigtriangleup$ & \cellcolor{gray!10}$ \bigtriangleup$ & Official\,(TensorFlow)\protect\footnotemark[34] \\\cline{3-10}
& & \emph{Meta-weight-net}\cite{shu2019meta} & \cellcolor{gray!10}$ \bigtriangleup$ & \cellcolor{blue!10}\cmark & \cellcolor{blue!10}\cmark & \cellcolor{red!10}\xmark &  \cellcolor{gray!10}$ \bigtriangleup$ & \cellcolor{gray!10}$ \bigtriangleup$ & Official\,(PyTorch)\protect\footnotemark[35] \\\cline{3-10}
& & {\emph{Data Coefficients}\cite{zhang2020distilling}} &  \cellcolor{blue!10}\cmark & \cellcolor{blue!10}\cmark & \cellcolor{blue!10}\cmark & \cellcolor{red!10}\xmark & \cellcolor{gray!10}$ \bigtriangleup$ & \cellcolor{gray!10}$ \bigtriangleup$ & Official\,(TensorFlow)\protect\footnotemark[36] \\\cline{3-10}%\\\midrule
& & \emph{Knowledge Distillation}\cite{li2017learning} & \cellcolor{blue!10}\cmark & \cellcolor{red!10}\xmark & \cellcolor{blue!10}\cmark & \cellcolor{red!10}\xmark &  \cellcolor{gray!10}$ \bigtriangleup$ & \cellcolor{gray!10}$ \bigtriangleup$ & N/A \\\cline{3-10}%\\\midrule
& & {\emph{MLC}\cite{zheng2021meta}} & \cellcolor{blue!10}\cmark & \cellcolor{blue!10}\cmark & \cellcolor{blue!10}\cmark & \cellcolor{red!10}\xmark &  \cellcolor{gray!10}$ \bigtriangleup$ & \cellcolor{blue!10}\cmark & Official\,(PyTorch)\protect\footnotemark[37] \\\midrule

\multirow{13}{*}{\vspace*{-0.45cm}\hspace*{0.13cm}\!\!\!\!\!\!\rotatebox[origin=c]{90}{\makecell[l]{Sample Selection\\\hspace*{0.5cm}(\textsection \ref{sec:sample_selection})}}\!\!\!\!\!} 
& \multirow{4}{*}{\makecell[l]{Multi-Network\\\hspace*{0.33cm}Learning}}  
& \emph{Decouple}\cite{malach2017decoupling} & \cellcolor{blue!10}\cmark & \cellcolor{blue!10}\cmark & \cellcolor{red!10}\xmark & \cellcolor{blue!10}\cmark & \cellcolor{red!10}\xmark & \cellcolor{gray!10}$ \bigtriangleup$ & Official\,(TensorFlow)\protect\footnotemark[38] \\\cline{3-10}
& & \emph{MentorNet}\cite{jiang2018mentornet} & \cellcolor{red!10}\xmark & \cellcolor{red!10}\xmark & \cellcolor{red!10}\xmark & \cellcolor{red!10}\xmark & \cellcolor{blue!10}\cmark & \cellcolor{gray!10}$ \bigtriangleup$ & Official\,(TensorFlow)\protect\footnotemark[39] \\\cline{3-10}
& & \emph{Co-teaching}\cite{han2018co} & \cellcolor{blue!10}\cmark & \cellcolor{blue!10}\cmark & \cellcolor{red!10}\xmark & \cellcolor{red!10}\xmark & \cellcolor{blue!10}\cmark & \cellcolor{gray!10}$ \bigtriangleup$ & Official\,(PyTorch)\protect\footnotemark[40] \\\cline{3-10}
& &  \emph{Co-teaching+}\cite{yu2019does} & \cellcolor{blue!10}\cmark & \cellcolor{blue!10}\cmark & \cellcolor{red!10}\xmark & \cellcolor{red!10}\xmark & \cellcolor{blue!10}\cmark & \cellcolor{gray!10}$ \bigtriangleup$ & Official\,(PyTorch)\protect\footnotemark[41] \\\cline{3-10}
& &  {\emph{JoCoR}\cite{wei2020combating}} & \cellcolor{blue!10}\cmark & \cellcolor{blue!10}\cmark & \cellcolor{red!10}\xmark & \cellcolor{red!10}\xmark & \cellcolor{blue!10}\cmark & \cellcolor{gray!10}$ \bigtriangleup$ & Official\,(PyTorch)\protect\footnotemark[42] \\\cmidrule{2-10}

& \multirow{6}{*}{\makecell[l]{Multi-Round\\\hspace*{0.23cm}Learning}}  
& \emph{ITLM}\cite{shen2019learning} & \cellcolor{blue!10}\cmark & \cellcolor{blue!10}\cmark & \cellcolor{red!10}\xmark & \cellcolor{red!10}\xmark & \cellcolor{blue!10}\cmark & \cellcolor{gray!10}$ \bigtriangleup$ & Official\,(GluonCV)\protect\footnotemark[43] \\\cline{3-10}
& & \emph{INCV}\cite{chen2019understanding} & \cellcolor{blue!10}\cmark &\cellcolor{blue!10}\cmark & \cellcolor{red!10}\xmark & \cellcolor{blue!10}\cmark & \cellcolor{blue!10}\cmark & \cellcolor{gray!10}$ \bigtriangleup$ & Official\,(Keras)\protect\footnotemark[44] \\\cline{3-10}
& & {\emph{O2U-Net}\cite{huang2019o2u}} & \cellcolor{blue!10}\cmark &\cellcolor{blue!10}\cmark & \cellcolor{red!10}\xmark & \cellcolor{red!10}\xmark & \cellcolor{blue!10}\cmark & \cellcolor{gray!10}$ \bigtriangleup$ & Unofficial\,(PyTorch)\protect\footnotemark[45] \\\cline{3-10}
& &  \emph{Iterative Detection}\cite{wang2018iterative} & \cellcolor{blue!10}\cmark & \cellcolor{blue!10}\cmark & \cellcolor{red!10}\xmark & \cellcolor{blue!10}\cmark & \cellcolor{blue!10}\cmark & \cellcolor{gray!10}$ \bigtriangleup$ & Official\,(Keras)\protect\footnotemark[46] \\\cline{3-10}
& &  {\emph{MORPH}\cite{song2020two}} & \cellcolor{blue!10}\cmark & \cellcolor{blue!10}\cmark & \cellcolor{red!10}\xmark & \cellcolor{blue!10}\cmark & \cellcolor{blue!10}\cmark & \cellcolor{gray!10}$ \bigtriangleup$ & N/A \\\cline{3-10}
& &  {\emph{TopoFilter}\cite{wu2020topological}} & \cellcolor{blue!10}\cmark & \cellcolor{blue!10}\cmark & \cellcolor{red!10}\xmark & \cellcolor{blue!10}\cmark & \cellcolor{blue!10}\cmark & \cellcolor{gray!10}$ \bigtriangleup$ & Official\,(PyTorch)\protect\footnotemark[47] \\\cline{3-10}
& &  {\emph{NGC}\cite{wu2021ngc}} & \cellcolor{blue!10}\cmark & \cellcolor{blue!10}\cmark & \cellcolor{blue!10}\cmark & \cellcolor{blue!10}\cmark & \cellcolor{blue!10}\cmark & \cellcolor{gray!10}$ \bigtriangleup$ & N/A\\\cmidrule{2-10}

& \multirow{3}{*}{\vspace*{-0.24cm}\makecell[l]{Hybrid Approach}}  
& \emph{SELFIE}\cite{song2019selfie} & \cellcolor{blue!10}\cmark & \cellcolor{blue!10}\cmark & \cellcolor{blue!10}\cmark & \cellcolor{red!10}\xmark & \cellcolor{blue!10}\cmark & \cellcolor{gray!10}$ \bigtriangleup$ & Official\,(TensorFlow)\protect\footnotemark[48] \\\cline{3-10}
& & \emph{SELF}\cite{nguyen2020self} & \cellcolor{blue!10}\cmark & \cellcolor{blue!10}\cmark & \cellcolor{blue!10}\cmark & \cellcolor{blue!10}\cmark & \cellcolor{blue!10}\cmark & \cellcolor{gray!10}$ \bigtriangleup$ & N/A \\\cline{3-10}
& & \emph{DivideMix}\cite{li2020dividemix} & \cellcolor{blue!10}\cmark & \cellcolor{blue!10}\cmark & \cellcolor{blue!10}\cmark & \cellcolor{blue!10}\cmark & \cellcolor{blue!10}\cmark & \cellcolor{gray!10}$ \bigtriangleup$ & Official\,(PyTorch)\protect\footnotemark[49] \\\cline{3-10}
& & {\emph{RoCL}\cite{zhou2021robust}} & \cellcolor{blue!10}\cmark & \cellcolor{blue!10}\cmark & \cellcolor{blue!10}\cmark & \cellcolor{blue!10}\cmark & \cellcolor{blue!10}\cmark & \cellcolor{gray!10}$ \bigtriangleup$ & N/A \\\bottomrule
\end{tabular}
\label{table:all_comparision}
\vspace*{-4cm}
\end{table*}

}

\clearpage
{
\footnotetext[1]{\url{https://github.com/endernewton/webly-supervised}}
\footnotetext[2]{\url{https://github.com/delchiaro/training-cnn-noisy-labels-keras}}
\footnotetext[3]{\url{https://github.com/ijindal/Noisy_Dropout_regularization}}
\footnotetext[4]{\url{https://github.com/udibr/noisy_labels}}
\footnotetext[5]{\url{https://github.com/Ryo-Ito/Noisy-Labels-Neural-Network}}
\footnotetext[6]{\url{https://github.com/Cysu/noisy_label}}
\footnotetext[7]{\url{https://github.com/bhanML/Masking}}
\footnotetext[8]{\url{https://github.com/pokaxpoka/RoGNoisyLabel}}
\footnotetext[9]{\url{https://github.com/sjenni/DeepBilevel}}
\footnotetext[10]{\url{https://rt416.github.io/pdf/trace_codes.pdf}}
\footnotetext[11]{\url{github.com/hendrycks/pre-training}}
\footnotetext[12]{\url{https://github.com/dmizr/phuber}}
\footnotetext[13]{\url{https://github.com/xiaoboxia/CDR}}
\footnotetext[14]{\url{https://github.com/hongxin001/ODNL?ref=pythonrepo.com}}
\footnotetext[15]{\url{https://https://github.com/sarathknv/adversarial-examples-pytorch}}
\footnotetext[16]{\url{https://github.com/CoinCheung/pytorch-loss}}
\footnotetext[17]{\url{https://github.com/facebookresearch/mixup-cifar10}}
\footnotetext[18]{\url{https://github.com/AlanChou/Truncated-Loss}}
\footnotetext[19]{\href{https://github.com/YisenWang/symmetric\_cross\_entropy\_for\_noisy_label}{https://github.com/YisenWang/symmetric\_cross\_entropy}}
\footnotetext[20]{\url{https://github.com/google/bi-tempered-loss}}
\footnotetext[21]{\url{https://github.com/HanxunH/Active-Passive-Losses}}
\footnotetext[22]{\url{https://github.com/giorgiop/loss-correction}}
\footnotetext[23]{\url{https://github.com/mmazeika/glc}}
\footnotetext[24]{\url{https://github.com/xiaoboxia/T-Revision}}
\footnotetext[25]{\href{https://github.com/xiaoboxia/Classification-with-noisy-labels-by-importance-reweighting}{https://github.com/xiaoboxia/Classification-with-noisy-labels}}

\newcolumntype{L}[1]{>{\centering\let\newline\\\arraybackslash\hspace{0pt}}m{#1}}
\newcolumntype{X}[1]{>{\centering\let\newline\\\arraybackslash\hspace{0pt}}p{#1}}
\begin{table*}[!htb]
\centering
\vspace*{-0.1cm}
\caption{Comparison of robust deep learning categories for overcoming noisy labels.}
\vspace*{-0.2cm}
\begin{tabular}{L{2.2cm} | L{2.85cm}|X{1.5cm}|X{1.55cm}|X{1.8cm}|X{1.65cm}|X{1.5cm}|X{1.6cm}}
\toprule
\multicolumn{2}{>{}c|}{\multirow{2}{*}{\textbf{Category}}} & \textbf{P1} & \textbf{P2} & \textbf{P3} & \textbf{P4} & \textbf{P5} & \textbf{P6} \\
\multicolumn{2}{>{}c|}{} & \!\!\!{Flexibility}\!\!\! & \!\!\!{No Pre-train}\!\!\! & \!\!\!{Full Exploration}\!\!\! & \!\!\!{No Supervision}\!\!\!  & \!\!\!{Heavy Noise}\!\!\! & \!\!\!{Complex Noise}\!\!\! \\\midrule
\multirow{2}{*}
{\makecell[l]{\!\!\!Robust Architecture\!\!\! \\\hspace*{0.5cm}(\textsection \ref{sec:robust_architecture})}}
& \!\!Noise Adaptation Layer & \cellcolor{gray!10}$ \bigtriangleup$ & \cellcolor{blue!10}\cmark & \cellcolor{blue!10}\cmark & \cellcolor{blue!10}\cmark  & \cellcolor{red!10}\xmark & \cellcolor{red!10}\xmark \\
\cline{2-8}
& \!\!Dedicated Architecture & \cellcolor{red!10}\xmark & \cellcolor{gray!10}$ \bigtriangleup$ & \cellcolor{blue!10}\cmark & \cellcolor{gray!10}$ \bigtriangleup$  & \cellcolor{gray!10}$ \bigtriangleup$ & \cellcolor{blue!10}\cmark\\ \midrule
\multirow{2}{*}
{\makecell[l]{\!\!\!\!\!\!Robust Regularization\!\!\!\!\!\! \\\hspace*{0.5cm}(\textsection \ref{sec:robust_regularization})}}
& \!\!Implicit Regularization & \cellcolor{blue!10}\cmark & \cellcolor{blue!10}\cmark & \cellcolor{blue!10}\cmark & \cellcolor{blue!10}\cmark  & \cellcolor{gray!10}$ \bigtriangleup$ & \cellcolor{gray!10}$ \bigtriangleup$ \\
\cline{2-8}
& \!\!Explicit Regularization & \cellcolor{blue!10}\cmark & \cellcolor{blue!10}\cmark & \cellcolor{blue!10}\cmark & \cellcolor{blue!10}\cmark  & \cellcolor{red!10}\xmark & \cellcolor{gray!10}$ \bigtriangleup$ \\ \midrule
%\multicolumn{2}{>{}c|}{Robust Regularization (\textsection \ref{sec:robust_regularization})}& \cellcolor{blue!10}\cmark & \cellcolor{blue!10}\cmark & \cellcolor{blue!10}\cmark & \cellcolor{blue!10}\cmark & \cellcolor{gray!10}$ \bigtriangleup$ & \cellcolor{gray!10}$ \bigtriangleup$ \\\hline
\multicolumn{2}{>{}c|}{\hspace*{0cm}Robust Loss Function (\textsection \ref{sec:robust_loss_function}) }& \cellcolor{blue!10}\cmark & \cellcolor{blue!10}\cmark & \cellcolor{blue!10}\cmark & \cellcolor{blue!10}\cmark  & \cellcolor{red!10}\xmark & \cellcolor{red!10}\xmark \\\midrule
\multirow{2}{*}
{\vspace*{-0.58cm}\!\makecell[l]{Loss Adjustment \\\hspace*{0.52cm}(\textsection \ref{sec:loss_adjustment})}}
& \!\!Loss Correction &\cellcolor{blue!10}\cmark & \cellcolor{red!10}\xmark & \cellcolor{blue!10}\cmark & \cellcolor{red!10}\xmark  & \cellcolor{red!10}\xmark &\cellcolor{red!10}\xmark \\
\cline{2-8}
& \!\!Loss Reweighting & \cellcolor{blue!10}\cmark & \cellcolor{blue!10}\cmark & \cellcolor{blue!10}\cmark & \cellcolor{blue!10}\cmark  & \cellcolor{red!10}\xmark & \cellcolor{gray!10}$ \bigtriangleup$\\
\cline{2-8}
& \!\!Label Refurbishment & \cellcolor{blue!10}\cmark & \cellcolor{blue!10}\cmark & \cellcolor{blue!10}\cmark & \cellcolor{blue!10}\cmark & \cellcolor{gray!10}$ \bigtriangleup$ & \cellcolor{gray!10}$ \bigtriangleup$\\\cline{2-8}
& \!\!Meta Learning & \cellcolor{blue!10}\cmark & \cellcolor{blue!10}\cmark & \cellcolor{blue!10}\cmark &  \cellcolor{red!10}\xmark & \cellcolor{gray!10}$ \bigtriangleup$  & \cellcolor{gray!10}$ \bigtriangleup$ \\ \midrule
{\vspace*{-0.55cm}\hspace*{0.075cm}\makecell[l]{Sample Selection\\\hspace*{0.55cm}(\textsection \ref{sec:sample_selection})}}
& \!\!Multi-Network Learning &\cellcolor{blue!10}\cmark &\cellcolor{blue!10}\cmark &  \cellcolor{red!10}\xmark & \cellcolor{red!10}\xmark  & \cellcolor{blue!10}\cmark  & \cellcolor{gray!10}$ \bigtriangleup$ \\
\cline{2-8}
& \!\!Multi-Round Learning & \cellcolor{blue!10}\cmark & \cellcolor{blue!10}\cmark & \cellcolor{red!10}\xmark & \cellcolor{blue!10}\cmark & \cellcolor{blue!10}\cmark & \cellcolor{gray!10}$ \bigtriangleup$\\\cline{2-8}
& \!\!Hybrid Approach & \cellcolor{blue!10}\cmark & \cellcolor{blue!10}\cmark & \cellcolor{blue!10}\cmark & \cellcolor{blue!10}\cmark & \cellcolor{blue!10}\cmark  & \cellcolor{gray!10}$ \bigtriangleup$\\\bottomrule
\end{tabular}
\label{table:direction_comparison}
\vspace*{-0.4cm}
\end{table*}

}

Meanwhile, \emph{SELFIE} \cite{song2019selfie} is a hybrid approach of sample selection and loss correction. The loss of refurbishable examples is corrected (i.e., loss correction) and then used together with that of small-loss examples (i.e., sample selection). Consequently, more training examples are considered for updating the DNN.
The \emph{curriculum loss\,(CL)} \cite{lyu2020curriculum} is combined with the robust loss function approach and used to extract the true-labeled examples from noisy data.% based on a manually specified selection threshold. 

\smallskip
\noindent \underline{Remark}:
Noise robustness is significantly improved by combining with other techniques. However, the hyperparameters introduced by these techniques render a DNN more susceptible to changes in data and noise types, and an increase in computational cost is inevitable

\section{Methodological Comparison}
\label{sec:comparison}

In this section, we compare the {$62$} deep learning methods for overcoming noisy labels introduced in Section \ref{sec:methodology} with respect to the following \emph{six} properties. When selecting the properties, we refer to the properties that are typically used to compare the performance of robust deep learning methods \cite{han2018co,song2019selfie}. To the best of our knowledge, this survey is the first to provide a systematic comparison of robust training methods. 
This comprehensive comparison will provide useful insights that can enlighten new future directions.
\vspace*{0.1cm}
\begin{itemize}[leftmargin=9pt]
\item \textbf{(P1)\,Flexibility:} With the rapid evolution of deep learning research, a number of new network architectures are constantly emerging and becoming available. Hence, the ability to support any type of architecture is important. \enquote{Flexibility} ensures that the proposed method can quickly adapt to the state-of-the-art architecture.
\vspace*{0.1cm}
\item \textbf{(P2)\,No Pre-training:} A typical approach to improve noise robustness is to use a pre-trained network; however, this incurs an additional computational cost to the learning process. \enquote{No Pre-training} ensures that the proposed method can be trained from scratch without any pre-training.
\vspace*{0.1cm}
\item \textbf{(P3)\,Full Exploration:} Excluding unreliable examples from the update is an effective method for robust deep learning; however, it eliminates hard but useful training examples as well. \enquote{Full Exploration} ensures that the proposed methods can use \emph{all} training examples without severe overfitting to false-labeled examples by adjusting their training losses or applying semi-supervised learning.
\vspace*{0.1cm}
\item \textbf{(P4)\,No Supervision:} Learning with supervision, such as a clean validation set or a known noise rate, is often impractical because they are difficult to obtain. Hence, such supervision had better be avoided to increase practicality in real-world scenarios. \enquote{No Supervision} ensures that the proposed methods can be trained without any supervision.
\vspace*{0.1cm}
\item \textbf{(P5)\,Heavy Noise:} In real-world noisy data, the noise rate can vary from light to heavy. Hence, learning methods should achieve consistent noise robustness with respect to the noise rate. \enquote{Heavy Noise} ensures that the proposed methods can combat even the heavy noise.
\end{itemize}

{
\footnotetext[26]{\url{https://github.com/songhwanjun/ActiveBias}}
\footnotetext[27]{\url{https://github.com/dr-darryl-wright/Noisy-Labels-with-Bootstrapping}}
\footnotetext[28]{\url{https://github.com/PaulAlbert31/LabelNoiseCorrection}}
\footnotetext[29]{\url{https://github.com/LayneH/self-adaptive-training}}
\footnotetext[30]{\url{https://github.com/xingjunm/dimensionality-driven-learning}}
\footnotetext[31]{\url{https://github.com/pingqingsheng/LRT}}
\footnotetext[32]{\url{https://github.com/chenpf1025/IDN}}
\footnotetext[33]{\url{https://github.com/krayush07/learn-by-weak-supervision}}
\footnotetext[34]{\url{https://github.com/uber-research/learning-to-reweight-examples}}
\footnotetext[35]{\url{https://github.com/xjtushujun/meta-weight-net}}
\footnotetext[36]{\url{https://github.com/google-research/google-research/tree/master/ieg}}
\footnotetext[37]{\url{https://aka.ms/MLC}}
\footnotetext[38]{\url{https://github.com/emalach/UpdateByDisagreement}}
\footnotetext[39]{\url{https://github.com/google/mentornet}}
\footnotetext[40]{\url{https://github.com/bhanML/Co-teaching}}
\footnotetext[41]{\url{https://github.com/bhanML/coteaching_plus}}
\footnotetext[42]{\url{https://github.com/hongxin001/JoCoR}}
\footnotetext[43]{\url{https://github.com/yanyao-shen/ITLM-simplecode}}
\footnotetext[44]{\url{https://github.com/chenpf1025/noisy_label_understanding_utilizing}}
\footnotetext[45]{\url{https://github.com/hjimce/O2U-Net}}
\footnotetext[46]{\url{https://github.com/YisenWang/Iterative_learning}}
\footnotetext[47]{\url{https://github.com/pxiangwu/TopoFilter}}
%\footnotetext[31]{\url{https://github.com/lpfgarcia/m2n}}
%\footnotetext[32]{\url{https://github.com/LiJunnan1992/MLNT}}
\footnotetext[48]{\url{https://github.com/kaist-dmlab/SELFIE}}
\footnotetext[49]{\url{https://github.com/LiJunnan1992/DivideMix}}
}

\begin{itemize}[leftmargin=9pt]
\item \textbf{(P6)\,Complex Noise:} The type of label noise significantly affects the performance of a learning method. To manage real-world noisy data, diverse types of label noise should be considered when designing a robust training method. \enquote{Complex Noise} ensures that the proposed method can combat even the complex label noise.
\end{itemize}

\vspace*{0.05cm}
Table \ref{table:all_comparision} shows a comparison of all robust deep learning methods, which are grouped according to the most appropriate category. In the first row, the aforementioned six properties are labeled as P1--P6, and the availability of open-source implementation is added in the last column. For each property, we assign \enquote{\cmark} if it is completely supported, \enquote{\xmark} if it is not supported, and \enquote{$\bigtriangleup$} if it is supported but not completely. 
%More specifically, \enquote{$\bigtriangleup$} is assigned to P1 if the method can be flexible but requires additional effort, 
%to P5 if the method can combat only moderate label noise, and to P6 if the method does not make a strict assumption about the noise type but without explicitly modeling instance-dependent noise. 
More specifically, \enquote{$\bigtriangleup$} is assigned to P1 if the method can be flexible but requires additional effort, to P5 if the method can combat only moderate label noise, {and to P6 if the method does not make a strict assumption about the noise type but without explicitly modeling instance-dependent noise. Thus, for P6, the method marked with \enquote{\xmark} only deals with the instance-independent noise, while the method marked with \enquote{\cmark} deals with both instance-independent and -dependent noises.} The remaining properties\,(i.e., P2, P3, and P4) are only assigned \enquote{\cmark} or \enquote{\xmark}. Regarding the implementation, we assign \enquote{N/A} if a publicly available source code is not available.

No existing method supports all the properties. Each method achieves noise robustness by supporting a different combination of the properties. The supported properties are similar among the methods of the same (sub-)category because those methods share the same methodological philosophy; however, they differ significantly depending on the (sub-)category. Therefore, we investigate the properties generally supported in each (sub-)category and summarize them in Table \ref{table:direction_comparison}. Here, the property of a (sub-)category is marked as the majority  of the belonging methods. If no clear trend is observed among those methods, then the property is marked \enquote{$\bigtriangleup$}.

%\noindent \circled{\color{white}5} \textit{Many label noise robust approach rely on efficient estimation of noise rate. There are in fact lot of papers on just noise rate estimation. I seriously feel that there should be a separate section on this issue and describe various approaches which can be used for noise rate estimation.} You are right, and this is a very good suggestion. Noise rate estimation can be done using the \emph{Gaussian mixture model} or \emph{cross validation}. We added Sec.~X-X to concentrate on the representative approaches (pp.\ x--x). 

\section{Noise Rate Estimation}
\label{sec:noise_rate}

{
The estimation of a noise rate is an imperative part of utilizing robust methods for better practical use, especially with the approaches belonging to the loss adjustment and sample selection. The estimated noise rate is widely used to reweight examples for a robust classifier \cite{zhang2018generalized, yao2020dual, liu2015classification} or to determine how many examples should be selected as clean ones \cite{han2018co, song2019selfie, chen2019understanding}. However, detailed analysis has yet to be performed properly, though many robust approaches highly rely on the accuracy of noise rate estimation. The noise rate can be estimated by exploiting the inferred noise transition matrix \cite{xia2019anchor, yao2020dual, li2021provably}, the Gaussian mixture model \cite{arazo2019unsupervised, pleiss2020detecting, song2020two}, or the cross-validation \cite{chen2019understanding, song2019selfie}.
}

\vspace*{-0.15cm}
\subsection{Noise Transition Matrix}

{
The noise transition matrix has been used to build a statistically consistent robust classifier because it represents the class posterior probabilities for noisy and clean data, as in Eq.\,\eqref{eq:label_transition_matrix}. The first method to estimate the noise rate is exploiting this noise transition matrix, which can be inferred or trained accurately by using perfectly clean examples, i.e., \emph{anchor points} \cite{liu2015classification, scott2015rate}; an example $x$ with its label $i$ is defined as an anchor point if $p(y=i|x)=1$ and $p(y=k|x)=0$ for $k \neq i$. Thus, let $\mathcal{A}_i$ be the set of anchor points with label $i$, then the element of the noise transition matrix $T_{ij}$ is estimated by \looseness=-1
\begin{equation}
\begin{split}
% &= \frac{1}{|\mathcal{A}_{i}|}\sum_{x\in\mathcal{A}_{i}} p(\tilde{y}=j|y=i)p(y=i|x) \\
\hat{T}_{ij} &= \frac{1}{|\mathcal{A}_{i}|}\sum_{x\in\mathcal{A}_{i}} \sum_{k=1}^{c} p(\tilde{y}=j|y=k)p(y=k|x) \\
&= \frac{1}{|\mathcal{A}_{i}|}\sum_{x\in\mathcal{A}_{i}} p(\tilde{y}=j|x; \Theta),  
%= \frac{1}{|\mathcal{A}_{i}|}\sum_{x\in\mathcal{A}_{i}} p(\tilde{y}=j|x; \Theta),  
\end{split}
\end{equation}
where $p(\tilde{y}=j|x; \Theta)$ is the noisy class posterior probability of the classifier trained on noisy training data for the anchor point $x$ (see the detailed proof in \cite{xia2019anchor, hendrycks2018using, yao2020dual}). Next, based on the inferred noise transition matrix, the noise rate of a balanced training data is obtained by averaging the label transition probabilities between classes,
\begin{equation}
\hat{\tau} = \frac{1}{c} \sum_{i=1}^{c} \sum_{j\neq i}^{c} p(\tilde{y}=j | {y}=i) = \frac{1}{c} \sum_{i=1}^{c} \sum_{j\neq i}^{c} \hat{T}_{ij}.
\end{equation}
However, since the anchor points are typically unknown in real-world data, they are identified from noisy training data by either theoretical derivations \cite{liu2015classification} or heuristics \cite{patrini2017making}. 
In addition, there have been recent efforts to learn the noise transition matrix without anchor points. \emph{T-Revision} \cite{xia2019anchor} initializes a transition matrix by exploiting the examples with high noisy class posterior probabilities and then refines the matrix by adding a slack variable. \emph{Dual-T} \cite{yao2020dual} introduces an intermediate class that factorizes the transition matrix into  two easy-to-estimate matrices for better accuracy. \emph{VolMinNet} \cite{li2021provably} realizes an end-to-end framework and relaxes the need for anchor points under the sufficiently scattered assumption.
}

{
\newcolumntype{L}[1]{>{\centering\let\newline\\\arraybackslash\hspace{0pt}}m{#1}}
\newcolumntype{X}[1]{>{\centering\let\newline\\\arraybackslash\hspace{0pt}}p{#1}}
\newcolumntype{Y}[1]{>{\let\newline\\\arraybackslash\hspace{1pt}}m{#1}}

\begin{table*}[t!]
\small
\caption{Summary of publicly available datasets used for studying label noise.}
\vspace*{-0.7cm}
\begin{center}
\begin{tabular}{L{2.0cm} | Y{3.4cm} |X{1.9cm}| X{1.9cm} |X{1.9cm} |X{1.9cm} |Y{2.1cm}}\toprule 
\multicolumn{2}{c|}{\textbf{Dataset}} & \textbf{\# Training} & \textbf{\# Validation} & \textbf{\# Testing} & \textbf{\# Classes} & \textbf{Noise Rate\,(\%)} \\\hline
\multirow{7}{*}{\hspace*{0.13cm}\!\!\!\!\!\!{\makecell[l]{Clean Data}}\!\!\!\!\!} 
& {MNIST} \cite{lecun1998mnist}\footnote[50] & {60K}& N/A  & {10K} & $10$ & $\approx0.0$ \\\cline{2-7}
& {Fashion-MNIST} \cite{xiao2017fashion}\footnote[51]\!\!\!\!\! & {60K}& N/A & {10K} & $10$ & $\approx0.0$ \\\cline{2-7}
& {CIFAR-10} \cite{krizhevsky2014cifar}\footnote[52] & {50K}& N/A & {10K} & $10$ & $\approx0.0$ \\\cline{2-7}
& {CIFAR-100} \cite{krizhevsky2014cifar}\footnote[52] & {50K}& N/A & {10K} & $100$ & $\approx0.0$ \\\cline{2-7}
& {SVHN} \cite{netzer2011reading}\footnote[53] & {73K}& N/A  & {26K} & $10$ & $\approx0.0$ \\\cline{2-7}
& {Tiny-ImageNet} \cite{karpathy2016cs231n}\footnote[55] & {100K}& {10K} & {10K} & $200$ & $\approx0.0$ \\\cline{2-7}
& {ImageNet} \cite{krizhevsky2012imagenet}\footnote[54] & {1.3M}& {50K}  & {50K} & $1000$ & $\approx0.0$ \\\midrule
\multirow{6}{*}{\hspace*{0.13cm}\!\!\!\!\!\!{\makecell[l]{Real-world\\\hspace*{-0.02cm}Noisy Data}}\!\!\!\!\!} 
& {ANIMAL-10N} \cite{song2019selfie}\footnote[56] & {50K}& N/A  & {5K} & $10$ & $\approx8.0$ \\\cline{2-7}
& {CIFAR-10N \cite{wei2021learning}}\footnote[57]  & {50K}& N/A  & {10K} & $10$ & $\approx9.0/18.0/40.2$\!\!\!\!\!\!\!  \\\cline{2-7}
& {CIFAR-100N \cite{wei2021learning}}\footnote[57]  & {50K}& N/A  & {10K} & $100$ & $\approx 25.6/40.2$ \\\cline{2-7}
& {Food-101N} \cite{lee2018cleannet}\footnote[58] & {310K}& {5K} & {25K} & $101$ & $\approx18.4$ \\\cline{2-7}
& {Clothing1M} \cite{xiao2015learning}\footnote[59] & {1M}& {14K} & {10K} & $14$ & $\approx38.5$ \\\cline{2-7}
& {WebVision} \cite{li2017webvision}\footnote[60] & {2.4M}& {50K} & {50K} & $1000$ & $\approx20.0$ \\\bottomrule
\end{tabular}
\end{center}
\label{table:summary_dataset}
\vspace*{-0.6cm}
\end{table*}
}

\vspace*{-0.15cm}
\subsection{Gaussian Mixture Model\,(GMM)}
\label{sec:gmm}

\begin{figure}[t!]
\begin{center}
\vspace*{0.1cm}
\includegraphics[width=8.8cm]{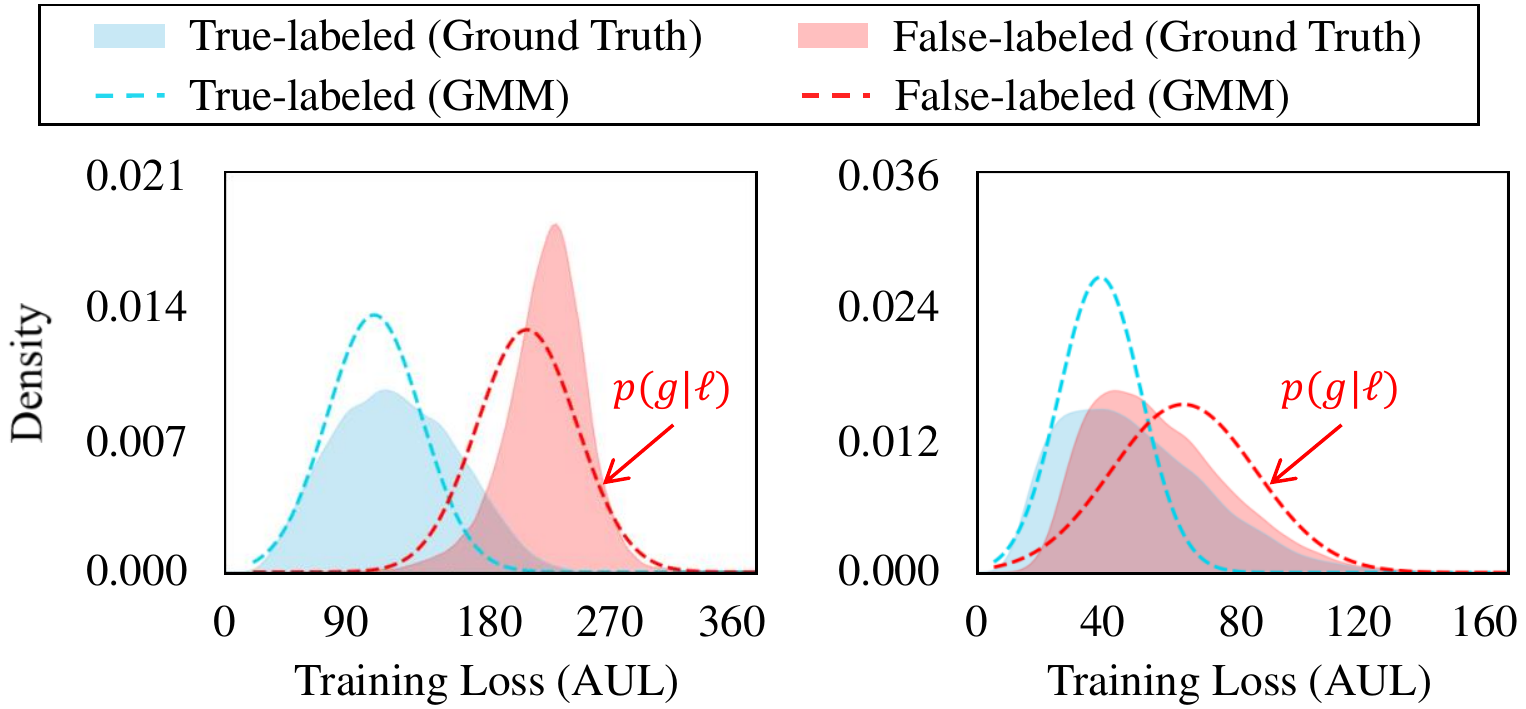}
\end{center}
\vspace*{-0.15cm}
\hspace*{1.3cm} \small{(a) Symmetric Noise.} \hspace*{1.2cm} \small{(b) Asymmetric Noise.} 
\vspace*{-0.15cm}
\caption{Training loss distributions of true-labeled and false-labeled examples using the ground-truth label and the GMM on CIFAR-100 data with two synthetic noises of $40\%$.}
\label{fig:loss_gmm}
\vspace*{-0.4cm}
\end{figure}

The second method is exploiting a one-dimensional and two-component GMM to model the loss distribution of true-labeled and false-labeled examples \cite{arazo2019unsupervised, pleiss2020detecting}. As shown in Figure \ref{fig:loss_gmm}, since the loss distribution tends to be \emph{bi-modal}, the two Gaussian components are fitted to the training loss by using the EM algorithm; the probability of an example being a false-labeled one is obtained through its posterior probability. Hence, the noise rate is estimated at each epoch $t$ by computing the expectation of the posterior probability for all training examples, \looseness=-1
\begin{equation}
\label{eq:naive_gmm}
\begin{gathered}
\hat{\tau} = \mathbb{E}_{(x,\tilde{y})\in\tilde{\mathcal{D}}}\Big[\,p\big(g\,|\,\ell\big(f(x;\Theta_{t}), \tilde{y}\big)\big)\,\Big],\\ 
%\hat{\tau} = \mathbb{E}_{(x,\tilde{y})\in\tilde{\mathcal{D}}}[p\big(g|AUL_{t}(x,\tilde{y})\big)],\\ 
%\text{where}~ \text{AUL}_{t}(x,\tilde{y})= \sum_{i=1}^{t}\ell(f(x;\theta_t),\tilde{y})
\end{gathered}
\end{equation}
where $g$ is the Gaussian component with a larger loss. However, Pleiss et al. \cite{pleiss2020detecting} recently pointed out that the training loss becomes less separable by the GMM as the training progresses, and thus proposed the \emph{area under the loss}\,(AUL) curve, which is the sum of the example's training losses obtained from all previous training epochs. Even after the loss signal decays in later epochs, the distributions remain separable. Therefore, the noise rate is finally estimated by
\begin{equation}
\label{eq:aul_gmm}
\begin{gathered}
\hat{\tau} = \mathbb{E}_{(x,\tilde{y})\in\tilde{\mathcal{D}}}\Big[\,p\big(g\,|\,{\rm AUL}_{t}(x,\tilde{y})\big)\,\Big],\\ 
\text{where}~ \text{AUL}_{t}(x,\tilde{y})= \sum_{i=1}^{t}\ell\big(f(x;\Theta_t),\tilde{y}\big).
\end{gathered}
\end{equation}

\vspace*{-0.4cm}
\subsection{Cross Validation}
\label{sec:cross_val}

The third method is estimating the noise rate by applying cross validation, which typically requires clean validation data \cite{song2019selfie, han2018co, yu2019does}. However, such clean validation data is hard to acquire in real-world applications. Thus, Chen et al. \cite{chen2019understanding} leveraged two randomly divided noisy training datasets for cross validation. Under the assumption that the two datasets share exactly the same noise transition matrix, the noise rate quantifies the test accuracy of DNNs that are respectively trained and tested on the two divided sets,
\begin{equation}
\label{eq:cross_val}
\!\!\!{\rm Test\,Accuracy} \!= \!\!\!\
\begin{cases}
(1 - \hat{\tau})^{2} +  \hat{\tau}^{2} / (c - 1) \!\!&\!\! \text{if symmetric}\!\!\!\!\\
(1 - \hat{\tau})^{2} + \hat{\tau}^{2}  \!\!& \!\!\text{if asymmetric}.\!\!\!\!
\end{cases}
\end{equation}
Therefore, the noise rate is estimated from the test accuracy obtained by cross validation.

\begin{comment}
\subsection{Comparison of Noise Rate Estimation}
\label{sec:nr_comp}

To compare the estimation performance of using the GMM and cross validation, we trained a WideResNet-16-8 for three benchmark datasets with varying noise rates. The results are plotted in Figure \ref{fig:noise_rate_symmetric}.  Generally, both methods performed well on the easy dataset\,(i.e., CIFAR-10), but their performance worsened as the training difficulty increased from CIFAR-10 to Tiny-ImageNet because the true-labeled but hard examples are not clearly distinguishable from the false-labeled ones. Nevertheless, the GMM method showed considerably better performance than the cross validation method even in the two difficult datasets, CIFAR-100 and Tiny-ImageNet. 
Overall, this empirical analysis will be helpful for practitioners or researchers who design robust algorithms for noisy labels.

\begin{figure}[!t]
\vspace*{0.0cm}
\begin{minipage}[t]{1.0\linewidth}
\begin{center}
\includegraphics[width=8.8cm]{figures/evaluation/noise_rate_symmetric.pdf}
\end{center}
\vspace*{-0.15cm}
\hspace*{2.1cm} \small{(a) GMM.} \hspace*{2.0cm} \small{(b) Cross Validation.}
\vspace*{-0.2cm}
\caption{Noise rate estimation using the Gaussian mixture model and cross validation when training a WideResNet-16-8 on three datasets with symmetric noise.}
\label{fig:noise_rate_symmetric}
\end{minipage}
\vspace*{-0.35cm}
\end{figure}

\end{comment}

\vspace*{-0.15cm}
\section{Experimental Design}

This section describes the typically used experimental design for comparing robust training methods in the presence of label noise. We introduce publicly available image datasets and then describe widely-used evaluation metrics.

\vspace*{-0.3cm}
\subsection{Publicly Available Datasets}

% Towards Robust Learning with Different Label Noise ..

To validate the robustness of the proposed algorithms, an image classification task was widely conducted on numerous image benchmark datasets. Table \ref{table:summary_dataset} summarizes popularly-used public benchmark datasets, which are classified into two categories: \emph{1)} a \enquote{clean dataset} that consists of mostly true-labeled examples annotated by human experts and \emph{2)} a \enquote{real-world noisy dataset} that comprises real-world noisy examples with varying numbers of false labels.

\vspace*{0.15cm}
\subsubsection{Clean Datasets}
According to the literature \cite{wang2018iterative, song2019selfie, li2020dividemix}, \emph{seven} clean datasets are widely used: MNIST\footnote[50]{\url{http://yann.lecun.com/exdb/mnist}}, classification of handwritten digits \cite{lecun1998mnist}; Fashion-MNIST\footnote[51]{\url{https://github.com/zalandoresearch/fashion-mnist}}, classification of various clothing \cite{xiao2017fashion}; CIFAR-10\footnote[52]{\url{https://www.cs.toronto.edu/~kriz/cifar.html}} and CIFAR-100\footnotemark[52], classification of a subset of $80$ million categorical images \cite{krizhevsky2014cifar}; SVHN\footnote[53]{\url{http://ufldl.stanford.edu/housenumbers}}, classification of house numbers in Google Street view images \cite{netzer2011reading};  ImageNet\footnote[54]{\url{http://www.image-net.org}} and Tiny-ImageNet\footnote[55]{\url{https://www.kaggle.com/c/tiny-imagenet}}, image database organized according to the WordNet hierarchy and its small subset \cite{krizhevsky2012imagenet, karpathy2016cs231n}. Because the labels in these datasets are almost all true-labeled, their labels in the training data should be artificially corrupted for the evaluation of synthetic noises, namely \emph{symmetric} noise and \emph{asymmetric} noise. 

\vspace*{0.15cm}
\subsubsection{Real-world Noisy Datasets} Unlike the clean datasets, real-world noisy datasets inherently contain many mislabeled examples annotated by non-experts. According to the literature \cite{li2017webvision, song2019selfie, lee2018cleannet, xiao2015learning}, \emph{six} real-world noisy datasets are widely used: ANIMAL-10N\footnote[56]{\url{https://dm.kaist.ac.kr/datasets/animal-10n}}, real-world noisy data of human-labeled online images for 10 confusing animals \cite{song2019selfie};
{CIFAR-10N\footnote[57]{\url{http://noisylabels.com/}} and CIFAR-100N\footnotemark[57], variations of CIFAR-10 and CIFAR-100 with human-annotated real-world noisy labels collected from Amazon’s Mechanical Turk \cite{wei2021learning}. They provide human labels with different noise rates, as shown in Table \ref{table:summary_dataset};}
Food-101N\footnote[58]{\url{https://kuanghuei.github.io/Food-101N}}, real-world noisy data of crawled food images annotated by their search keywords in the Food-101 taxonomy \cite{bossard2014food, lee2018cleannet}; Clothing1M\footnote[59]{\url{https://www.floydhub.com/lukasmyth/datasets/clothing1m}}, real-world noisy data of large-scale crawled clothing images from several online shopping websites \cite{xiao2015learning}; WebVision\footnote[60]{\url{https://data.vision.ee.ethz.ch/cvl/webvision/download.html}}, real-world noisy data of large-scale web images crawled from  Flickr  and Google Images search \cite{li2017webvision}. To support sophisticated evaluation, most real-world noisy datasets contain their own clean validation set and provide the estimated noise rate of their training set. \looseness=-1

% synthetic noise
% pseudo noise?
% real-wordl?
%\vspace*{-0.3cm}
\subsection{Evaluation Metrics}

A typical metric to assess the robustness of a particular method is the prediction accuracy for unbiased and clean examples that are not used in training. The prediction accuracy degrades significantly if the DNN overfits to false-labeled examples \cite{zhang2016understanding}. Hence, \emph{test accuracy} has generally been adopted for evaluation \cite{frenay2013classification}. For a test set $\mathcal{T}=\{(x_i,y_i)\}_{i=1}^{|\mathcal{T}|}$, let $\hat{y}_i$ be the predicted label of the $i$-th example in $\mathcal{T}$. Subsequently, the test accuracy is formalized by
\begin{equation}
\label{eq:test_accuracy}
\text{Test Accuracy} = \frac{|\{(x_i,y_i)\in\mathcal{T}:\hat{y}_i=y_i\}|}{|\mathcal{T}|}.
\end{equation}
If the test data are not available, \emph{validation accuracy} can be used by replacing $\mathcal{T}$ in Eq.\,\eqref{eq:test_accuracy} with validation data $\mathcal{V}=\{(x_i,y_i)\}_{i=1}^{|\mathcal{V}|}$ as an alternative,
\begin{equation}
\label{eq:validation_accuracy}
\text{Validation Accuracy} = \frac{|\{(x_i,y_i)\in\mathcal{V}:\hat{y}_i=y_i\}|}{|\mathcal{V}|}.
\end{equation}

Furthermore, if the specified method belongs to the \enquote{sample selection} category, \emph{label precision} and \emph{label recall} \cite{han2018co, chen2019understanding} can be used as the metrics,

\noindent
\begin{equation}
\label{eq:label_precision}
\begin{gathered}
\text{Label Precision} = \frac{|\{(x_i,\tilde{y}_i)\in\mathcal{S}_t: \tilde{y}_i = y_i\}|}{|\mathcal{S}_t|},\\
\text{Label Recall} = \frac{|\{(x_i,\tilde{y}_i)\in\mathcal{S}_t: \tilde{y}_i = y_i\}|}{|\{(x_i,\tilde{y}_i)\in\mathcal{B}_t: \tilde{y}_i = y_i\}|},
\end{gathered}
\end{equation}
where $\mathcal{S}_t$ is the set of selected clean examples in a mini-batch $\mathcal{B}_t$. The two metrics are performance indicators for the examples selected from the mini-batch as true-labeled ones \cite{han2018co}. \looseness=-1

Meanwhile, if the specified method belongs to the \enquote{label refurbishment} category, \emph{correction error} \cite{song2019selfie} can be used as an indicator of how many examples are incorrectly refurbished,
\begin{equation}
\label{eq:correction_error}
\!\text{Correction Error} = \frac{|\{x_i\!\in\!\mathcal{R}: \text{argmax}(y_{i}^{refurb}) \neq y_i\}|}{|\mathcal{R}|},
\end{equation}
where $\mathcal{R}$ is the set of examples whose labels are refurbished by Eq.\,(\ref{eq:label_correction}) and $y_i^{refurb}$ is the refurbished label of the $i$-th examples in $\mathcal{R}$. 

\vspace*{-0.12cm}
\section{Future Research Directions}
%\vspace*{-0.05cm}

With recent efforts in the machine learning community, the robustness of DNNs becomes evolving in several directions. Thus, the existing approaches covered in our survey face a variety of future challenges. This section provides discussion for future research that can facilitate and envision the development of deep learning in the label noise area. \looseness=-1

%This section presents a few challenging but interesting future research directions. 

\vspace*{-0.22cm}
\subsection{{Instance-dependent Label Noise}}
%\vspace*{-0.05cm}

{
Existing theoretical and empirical studies for \emph{robust loss function} and \emph{loss correction} are largely built upon the instance-independent noise assumption that the label noise is independent of input features \cite{sukhbaatar2014training, bekker2016training, xia2019anchor, yao2020dual}. However, this assumption may not be a good approximation of the real-world label noise. In particular, Chen et al. \cite{chen2021beyond} conducted a theoretical hypothesis testing\footnote[61]{In Clothing1M, the result showed that the instance-independent noise happens with probability lower than $10^{-21250}$, which is statistically impossible.} using a popular real-world dataset, Clothing1M, and proved that its label noise is statistically different from the instance-independent noise. This testing confirms that the label noise should depend on the instance. \looseness=-1

Conversely, most methods for the other direction (especially, \emph{sample selection}) work well even under the instance-dependent label noise in general since they do not rely on the assumption. Nevertheless, Song et al. \cite{song2019prestopping} pointed out that their performance could considerably worsen in the instance-dependent\,(or real-world) noise compared to symmetric noise due to the confusion between true-labeled and false-labeled examples. The loss distribution of true-labeled examples heavily overlaps that of false-labeled samples in the asymmetric noise, which is similar to the real-world noise, in Figure \ref{fig:loss_distribution}(b). Thus, identifying clean examples becomes more challenging when dealing with the instance-dependent label noise.%, though they are not directly affected by the noise type.

Beyond the instance-independent label noise, there have been a few recent studies for the instance-dependent label noise. Mostly, they only focus on a binary classification task \cite{menon2018learning, bootkrajang2020towards} or a restricted small-scale machine learning model such as logistic regression \cite{cheng2020learning}. Therefore, learning with the instance-dependent label noise is an important topic that deserves more research attention.} %for better practical use of deep learning.

\vspace*{-0.32cm}

\subsection{{Multi-label Data with Label Noise}}
%\vspace*{-0.05cm}

{
Most of the existing methods are applicable only for a \emph{single-label} multi-class classification problem, where each data example is assumed to have only one true label. However, in the case of \emph{multi-label} learning, each data example can be associated with a set of multiple true class labels. In music categorization, each music can belong to multiple categories \cite{tsoumakas2007multi}. In semantic scene classification, each scene may belong to multiple scene classes \cite{boutell2004learning}. Thus, contrary to the single-label setup, the multi-label classifier aims to predict a set of target objects simultaneously. 
In this setup, a multi-label dataset of millions of examples are reported to contain over $26.6\%$ false-positive labels as well as a significant number of omitted labels \cite{krasin2017openimages}. 

Even worse, the difference in occurrence between classes makes this problem more challenging; some minor class labels occur less in training data than other major class labels. Considering such aspects that can arise in multi-label classification, the simple extension of existing methods may not learn the proper correlations among multiple labels. Therefore, learning from noisy labels with multi-label data is another important topic for future research. We refer the readers to a recent study \cite{zhao2021evaluating} that discusses the evaluation of multi-label classifiers trained with noisy labels.
}

% pretraining. 
\vspace*{-0.12cm}
\subsection{{Class Imbalance Data with Label Noise}}

{
The \emph{class imbalance} in training data is commonly observed, where a few classes account for most of the data. Especially when working with large data in many real-world applications, this problem becomes more severe and is often associated with the problem of noisy labels \cite{johnson2019survey}.
%As data follows skewed distributions with a long-tail configuration, the label noise and class imbalance problems occur simultaneously in many real-world applications \cite{johnson2019survey}.
Nevertheless, to ease the label noise problem, it is commonly assumed that training examples are equally distributed over all class labels in the training data. This assumption is quite strong when collecting large-scale data, and thus we need to consider a more realistic scenario in which the two problems coexist. \looseness=-1

Most of the existing robust methods may not work well with the class imbalance, especially when they rely on the learning dynamics of DNNs, e.g., the small-loss trick or memorization effect. Under the existence of the class imbalance, the training model converges to major classes faster than minor classes such that most examples in the major class exhibit small losses\,(i.e., early memorization). That is, there is a risk of discarding most examples in the minor class. Furthermore, in terms of example importance, high-loss examples are commonly favored for the class imbalance problem \cite{shu2019meta}, while small-loss examples are favored for the label noise problem. This conceptual contradiction hinders the applicability of the existing methods that neglect the class imbalance. Therefore, these two problems should be considered simultaneously to deal with more general situations.}

\vspace*{-0.12cm}
\subsection{{Robust and Fair Training}}

{
Machine learning classifiers can perpetuate and amplify the existing systemic injustices in society \cite{hardt2016equality}. Hence, fairness is becoming another important topic. Traditionally, robust training and fair training have been studied by separate communities; robust training with noisy labels has mostly focused on combating label noise without regarding data bias \cite{frenay2013classification, han2020survey}, whereas fair training has focused on dealing with data bias, not necessarily noise \cite{hardt2016equality, jiang2020identifying}. However, noisy labels and data bias, in fact, coexist in real-world data. Satisfying both robustness and fairness is more realistic but challenging because the bias in data is pertinent to label noise. 

In general, many fairness criteria are group-based, where a target metric is equalized or enforced over subpopulations in the data, also known as \emph{protected groups} such as race or gender \cite{hardt2016equality}. Accordingly, the goal of fair training is building a model that satisfies such fairness criteria for the \emph{true} protected groups. However, if the \emph {noisy} protection group is involved, such fairness criteria cannot be directly applied. Recently, mostly after 2020, a few pioneering studies have emerged to consider both robustness and fairness objectives at the same time under the binary classification setting \cite{wang2020robust, wang2021fair}. Therefore, more research attention is needed for the convergence of robust training and fair training.
}

\vspace*{-0.12cm}
%\subsection{\!\!\!{\color{Rall}\Rac Connection between Input Perturbation and Label Noise\!\!\!\!\!\!}}
\subsection{{Connection with Input Perturbation}}

{
There has been a lot of research on the robustness of deep learning under input perturbation, mainly in the field of adversarial training where the  {input feature} is maliciously perturbed to distort the output of the DNN \cite{dohmatob2018limitations, mahloujifar2019curse}. Although learning with noisy labels and learning with noisy inputs have been regarded as separate research fields, their goals are similar in that they learn noise-robust representations from noisy data. Based on this common point of view, a few recent studies have investigated the interaction of adversarial training with noisy labels \cite{uesato2019labels, damodaran2020wasserstein, zhu2021understanding}.

Interestingly, it was turned out that adversarial training makes DNNs robust to label noise \cite{uesato2019labels}. Based on this finding, Damodaran et al. \cite{damodaran2020wasserstein} proposed a new regularization term, called Wasserstein adversarial regularization, to address the problem of learning with noisy labels. Zhu et al. \cite{zhu2021understanding} proposed to use the number of projected gradient descent steps as a new criterion for sample selection such that clean examples are filtered out from noisy data. These approaches are regarded as a new perspective on label noise compared to traditional work. Therefore, understanding the connection between input perturbation and label noise could be another future topic for better representation learning toward robustness. 
}

\vspace*{-0.12cm}
\subsection{{Efficient Learning Pipeline}}

{
{The efficiency of the learning pipeline is another important aspect to design deep learning approaches.} However, for robust deep learning, most studies have neglected the efficiency of the algorithm because their main goal is to improve the robustness to label noise. For example, maintaining multiple DNNs or training a DNN in multiple rounds is frequently used, but these approaches significantly degrade the efficiency of the learning pipeline. On ther other hand, the need for more efficient algorithms is increasing owing to the rapid increase in the amount of available data \cite{nguyen2019machine}. 

According to our literature survey, most work did not even report the efficiency\,(or time complexity) of their approaches. However, it is evident that saving the training time is helpful under the restricted budget for computation.
Therefore, enhancing the efficiency will significantly increase the usability of robust deep learning in the big data era.
}

\section{Conclusion}

DNNs easily overfit to false labels owing to their high capacity in totally memorizing all noisy training samples. This overfitting issue still remains even with various conventional regularization techniques, such as dropout and batch normalization, thereby significantly decreasing their generalization performance. Even worse, in real-world applications, the difficulty in labeling renders the overfitting issue more severe. Therefore, learning from noisy labels has recently become one of the most active research topics. 

In this survey, we presented a comprehensive understanding of modern deep learning methods to address the negative consequences of learning from noisy labels. All the methods were grouped into five categories according to their underlying strategies and described along with their methodological weaknesses. Furthermore, a systematic comparison was conducted using six popular properties used for evaluation in the recent literature. According to the comparison results, there is no ideal method that supports all the required properties; the supported properties varied depending on the category to which each method belonged. Several experimental guidelines were also discussed, including {noise rate estimation,} publicly available datasets, and evaluation metrics. Finally, we provided insights and directions for future research in this domain.


%\bibliography{6-reference.bib}
%\bibliographystyle{IEEEtran}

\section*{Acknowledgements}
This work was supported by Institute of Information \& Communications Technology Planning \& Evaluation\,(IITP) grant funded by the Korea government\,(MSIT) (No. 2020-0-00862, DB4DL: High-Usability and Performance In-Memory Distributed DBMS for Deep Learning).

%\input{0-paper.bbl}


%\end{comment}

%\newpage


%\vfill


\end{document}